\documentclass[smallcondensed]{svjour3}  

\smartqed  

\usepackage{amsmath,amssymb,amsfonts}

\DeclareMathOperator*{\argmin}{arg\,min}
\usepackage{units} 
\usepackage{bm}
\usepackage{float}
\usepackage[font=small,skip=7pt]{caption}
\usepackage[linesnumbered,noend,ruled,vlined]{algorithm2e}
\usepackage{hyperref}
\usepackage{microtype}
\usepackage{multirow}
\usepackage{graphicx}
\usepackage[table,xcdraw]{xcolor}
\usepackage{epsfig}
\usepackage{rotating}
\SetKwInput{KwInput}{Input}                
\SetKwInput{KwOutput}{Output}              
\newlength\fwidth
\usepackage{mathtools}

\usepackage{xcolor}
\usepackage{siunitx}
\usepackage{multirow}
\usepackage{epstopdf}
\usepackage{amsmath} 
\usepackage{amssymb}  
\usepackage{mathtools}
\usepackage[T1]{fontenc}
\DeclareGraphicsExtensions{.pdf,.eps}
\usepackage{textcomp}
\usepackage{subfigure}
\usepackage[ISO]{diffcoeff}
\usepackage{caption}

\def\BibTeX{{\rm B\kern-.05em{\sc i\kern-.025em b}\kern-.08em
    T\kern-.1667em\lower.7ex\hbox{E}\kern-.125emX}}

\usepackage[acronym,nomain,nonumberlist]{glossaries}
\newacronym{mav}{MAV}{Micro Aerial Vehicle}
\newacronym{nmpc}{NMPC}{Nonlinear Model Predictive Control}
\newacronym{mpc}{MPC}{Model Predictive Control}
\newacronym{panoc}{PANOC}{Proximal Averaged Newton-type method for Optimal Control}
\newacronym{gps}{GPS}{Global Positioning System}
\newacronym{cnn}{CNN}{Convolutional Neural Network}
\newacronym{uwb}{UWB}{Ultra-Wide Band}
\newacronym{mbp}{MBP}{Motion Primitives-Based Path Planner}
\newacronym{compra}{COMPRA}{Compact Reactive Autonomy}
\newacronym{sar}{SAR}{Search-and-Rescue}
\newacronym{apf}{APF}{Artificial Potential Field}
\newacronym{dphr}{DPHR}{Deepest-point Heading Regulation}

\usepackage{url}

\begin{document}

\title{\LARGE \bf REF: A Rapid Exploration Framework for Deploying Autonomous MAVs in Unknown Environments}
\titlerunning{Rapid SubT exploration framework}        

\author{Akash Patel$^1$ \and Bj\"orn Lindqvist$^1$ \and Christoforos Kanellakis$^1$ \and Ali-akbar Agha-mohammadi$^2$ \and George Nikolakopoulos$^1$
}


\institute{Akash Patel \at
              \email{akash.patel@ltu.se}           
           \and
           Bj\"orn Lindqvist \at
              \email{bjorn.lindqvist@ltu.se}           
           \and
          Christoforos Kanellakis \at
              \email{christoforos.kanellakis@ltu.se}           
            \and
           Ali-akbar Agha-mohammadi \at
              \email{aliakbar.aghamohammadi@jpl.nasa.gov}  
           \and
           George Nikolakopoulos \at
              \email{george.nikolakopoulos@ltu.se}           
\and
$^1$ The authors are with the Robotics and Artificial Intelligence Team, Department of Computer, Electrical and Space Engineering, Lule\r{a} University of Technology, Lule\r{a} SE-97187, Sweden. 
\and
$^{2}$The author is with Jet Propulsion Laboratory California Institute of Technology Pasadena, CA, 91109.
}

\date{Received: / Accepted: date}

\captionsetup{font=footnotesize}
\maketitle
\begin{abstract}\label{sec:abstract}
Exploration and mapping of unknown environments is a fundamental task in applications for autonomous robots. In this article, we present a complete framework for deploying Micro Aerial Vehicles (MAVs) in autonomous exploration missions in unknown subterranean areas. The main motive of exploration algorithms is to depict the next best frontier for the robot such that new ground can be covered in a fast, safe yet efficient manner. The proposed framework uses a novel frontier selection method that also contributes to the safe navigation of autonomous robots in obstructed areas such as subterranean caves, mines, and urban areas. The framework presented in this work bifurcates the exploration problem in local and global exploration. The proposed exploration framework is also adaptable according to computational resources available onboard the robot which means the trade-off between the speed of exploration and the quality of the map can be made. Such capability allows the proposed framework to be deployed in a subterranean exploration, mapping as well as in fast search and rescue scenarios. The performance of the proposed framework is evaluated in detailed simulation studies with comparisons made against a high-level exploration-planning framework developed for the DARPA Sub-T challenge as it will be presented in this article.

\footnote{The video link of this work can be found at \url{https://youtu.be/nmN0Xy6EqLM}}
\end{abstract}
\glsresetall 
%
\section{Introduction and Background}\label{sec:intro}

Rapid exploration and mapping of unknown subterranean environments has become significant interest in the field of autonomous deployment of robots. MAVs have the potential in being a viable solution in terms of mines inspection~\cite{kanellakis2018towards}, exploration and mapping~\cite{lindqvist2021exploration}, \cite{lindqvist2021compra}, \cite{Patel2022fast} and inspection of infrastructures~\cite{mansouri2018cooperative} due to their high degree of freedom and fast traversability. The applications of MAVs have also been discussed in developing next generation rotor crafts for mars exploration in \cite{patel2021design} and \cite{patel2021mcoax}. Deploying MAVs for exploration and mapping of dark, dusty and hostile mines and caving systems is particularly challenging because at the beginning of the exploration process, the environment is completely unknown for navigation. In order to map surrounding for safe navigation in such environments, vision-only based navigation techniques are insufficient \cite{ozaslan2017autonomous}. The unstructured and rocky environment of mines and caves is a major challenge that contribute in uncertainty in sensor measurements~\cite{agha2021nebula}. In order to explore and map such environments, the crucial requirements for autonomous navigation problem are: a) detecting the frontiers, b) selecting the optimal frontier, and c) safe navigation to such frontier in order to successfully build a map of the environment. In order to safely navigate in an unknown environment, it is crucial that the MAV is backed by a sophisticated on board autonomy for Guidance, Navigation and Control (GNC). 

\begin{figure}[h!]
  \centering
  \captionsetup{justification=centering}
    \includegraphics[width=\textwidth]{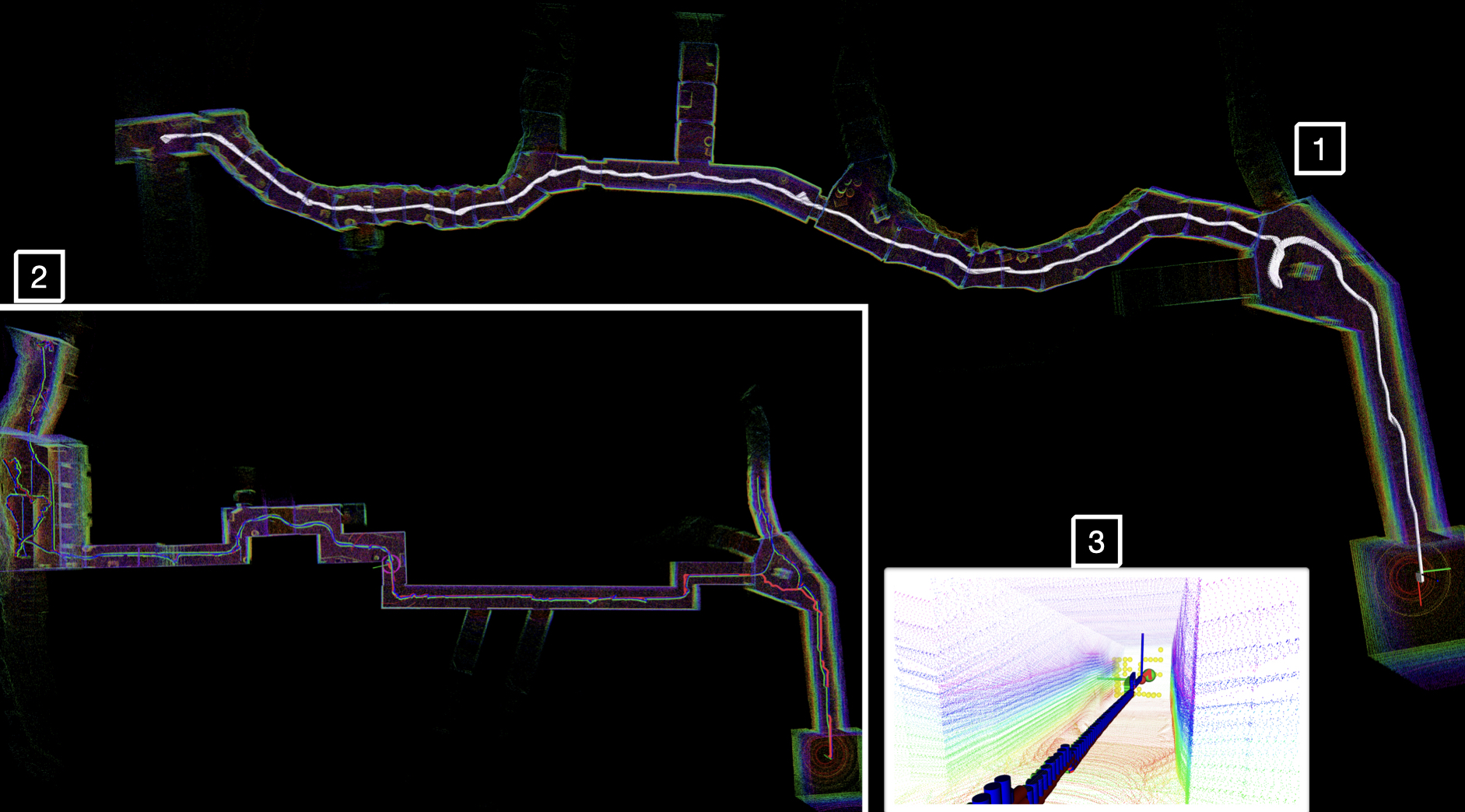}
    \caption{DARPA Sub-T world : Exploration instance. (1) Rapid local exploration behaviour (2) local exploration in very narrow as well as wide cave-void like areas (3) Safe Next Best Frontier (NBF) in obstructed narrow tunnels}
  \label{fig:snap1}
\end{figure}

\begin{figure}[h!]
  \centering
  \captionsetup{justification=centering}
    \includegraphics[width=\textwidth]{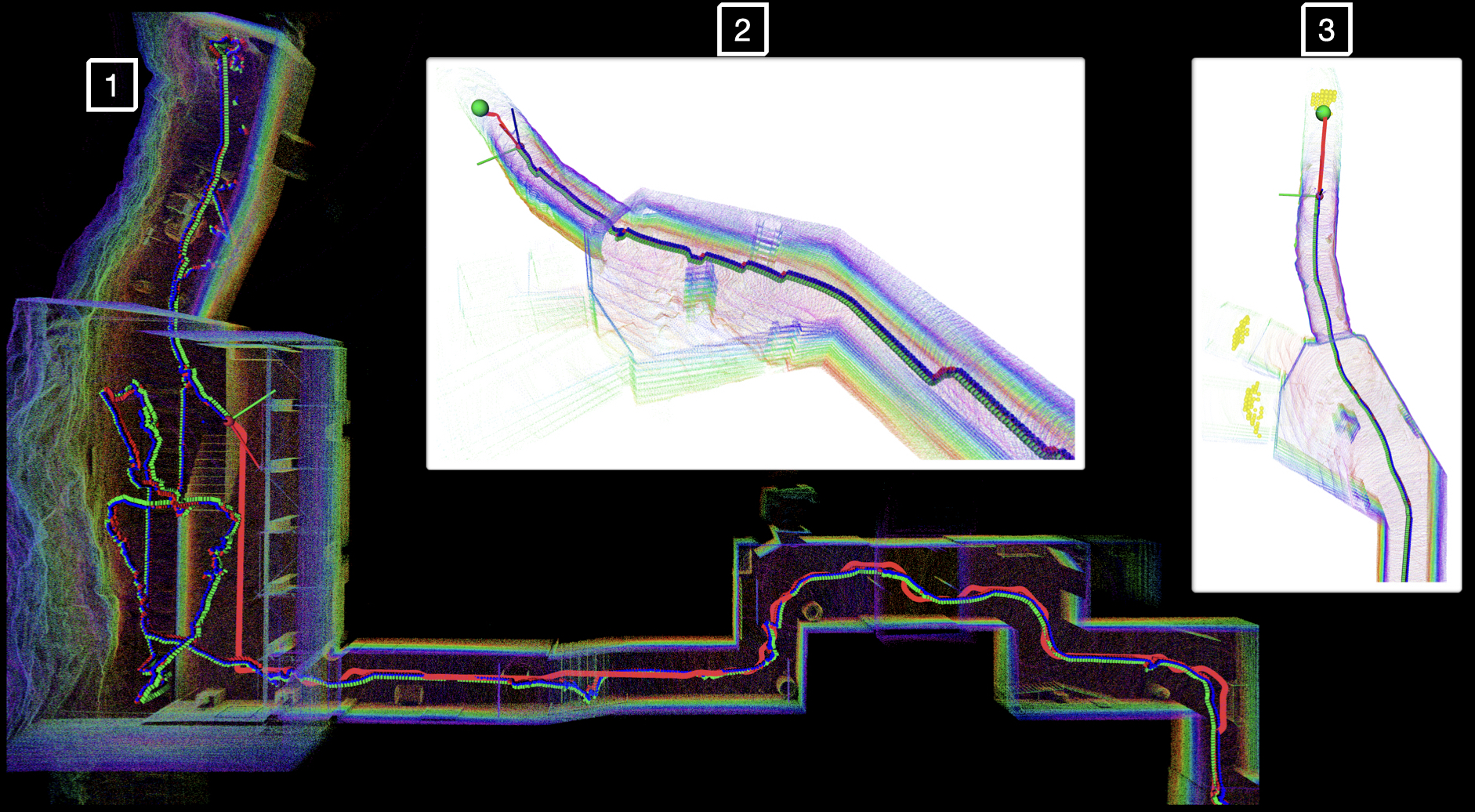}
    \caption{Exploration behaviour using the proposed framework in multiple exploration scenarios in DARPA Sub-T virtual world}
  \label{fig:snap2}
\end{figure}

In \autoref{fig:snap1} and \autoref{fig:snap2} exploration instances of the proposed method is shown in different environments. The capability of the proposed method to handle exploration of narrow passages as well as wide tall void like structures is evident in \autoref{fig:snap2}.

\begin{figure}[h!]
  \centering
  \captionsetup{justification=centering}
    \includegraphics[width=0.8\linewidth]{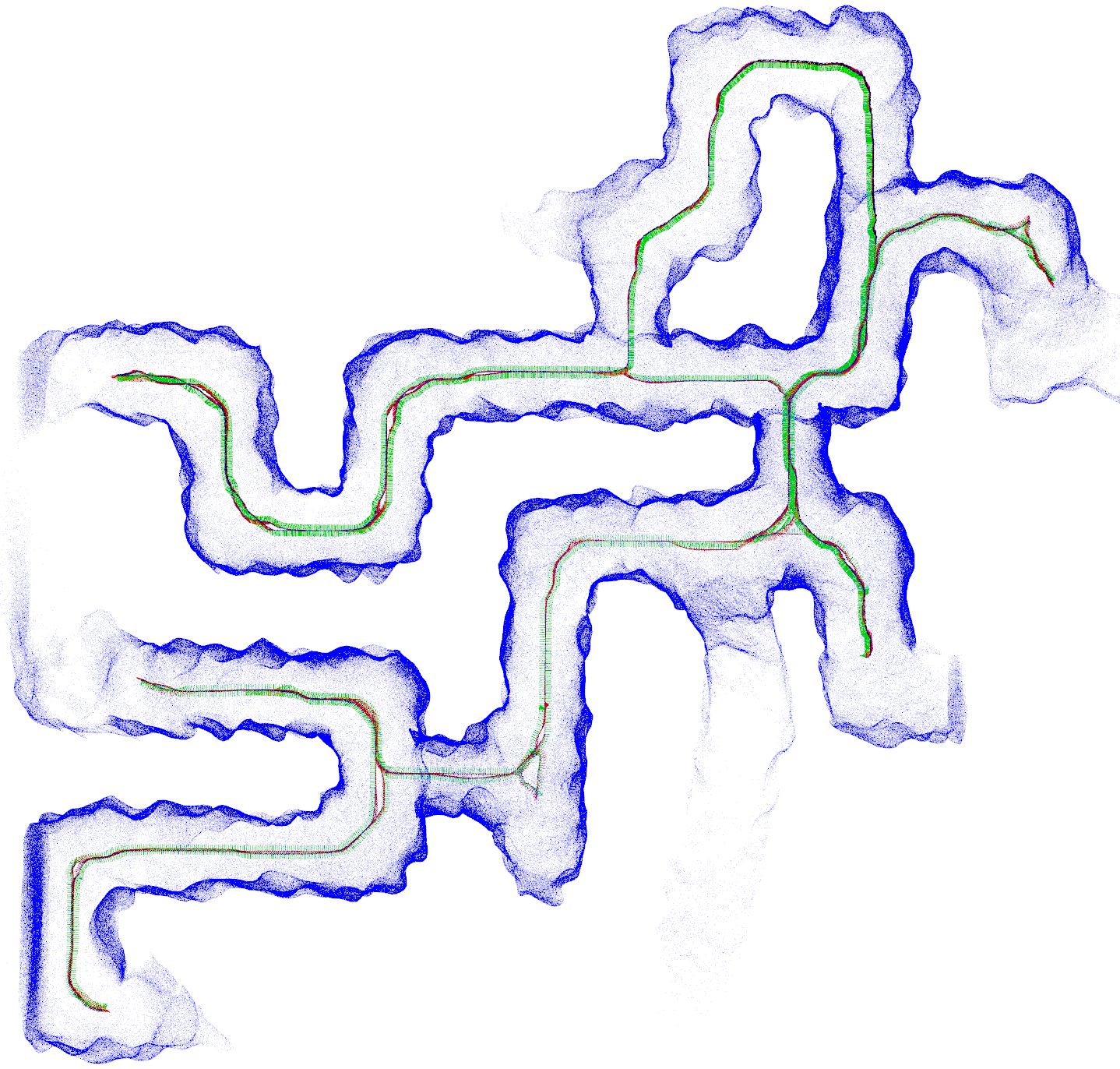}
    \caption{MAV trajectory while exploring a wide-large cave environment using the proposed approach}
  \label{fig:reconstruction}
\end{figure}

The framework introduced in this work selects optimal frontiers based on the idea of continuing the exploration in one direction until there is no new potential information to gain in the particular direction. Planning a safe path to such selected frontiers is crucial when exploring a large environment. The path planning method used in this work takes into account the safety margin of such paths based on the size of the MAV and its ability to traverse through the obstructed areas. The MAVs are also constrained in terms of their limited time of flight. Therefore the proposed framework also accounts for cost based frontier selection while evaluating next optimal area to visit. The proposed framework also complements the idea of efficiently utilising the resources of the vehicle by rapid yet safe navigation. This work presents a rapid exploration framework for safe autonomous navigation of MAVs in caves. The point cloud map of the explored virtual cave environment with the MAV's trajectory is presented in \autoref{fig:reconstruction}. 

\subsection{Related Works}\label{sec:related_works}
In the original work of frontier based exploration~\cite{yamauchi1997frontier}, the points lying at the boundary between known (free) space and unknown space are defined as frontier points. In~\cite{yamauchi1997frontier} a closest frontier from the robots position is selected to move to such that the boundary at which frontiers lie will also progress towards more unexplored space. The same approach was also extended for the case of multiple robots, as presented in~\cite{yamauchi1998frontier}. In~\cite{holz2010evaluating} and~\cite{julia2012comparison} frontier based exploration strategies are studied extensively for comparison against different exploration approaches. A 3D Frontier Based Exploration Tool (FBET) for aerial vehicles is presented in \cite{zhu20153d}. The FBET framework uses similar approach to \cite{yamauchi1997frontier} for frontiers generation and the generated frontier are clustered for selection of candidate frontier goal based on cost function that takes into account the cost of moving to such goal point. A Stochastic Differential Equation (SDE) based exploration approach is presented in \cite{shen2012autonomous}. In the SDE based exploration strategy the authors consider simulating expansion of system of particles with Newtonian dynamics for evolution of SDE. In \cite{shen2012autonomous} the authors consider the region showing significant expansion of particles as a region that would lead the MAV to more unexplored space. In \cite{fraundorfer2012vision} A vision based exploration-mapping problem solving technique is presented that also utilizes MAV to navigate in unexplored areas using continuously updating frontiers. Exploration of unknown environments are also extended to legged or ground robots. Probabilistic Local and Global Reasoning on Information roadMaps (PLGRIM) as presented in \cite{kim2021plgrim}, discusses a hierarchical value learning strategy for autonomous exploration of large subterranean environments. The methodology presented in~\cite{kim2021plgrim} uses a hierarchical learning to address local and global exploration of large scale environments while focusing on near optimal coverage plans. A Frontloaded Information Gain Orienteering Problem (FIG-OP) based strategy is presented in~\cite{peltzer2022fig} that uses topological maps to plan exploration paths in fixed time budget exploration scenario. The method presented in~\cite{peltzer2022fig} is tested with ground robots in multi-kilometers subterranean environment targeted at time constrained exploration missions. 
Separated from frontier-selection methods are the methods with integrated exploration behavior in the path planning problem, often based on trying to plan a path in order to maximize the information gain, while minimizing distance travelled or similar metrics. These planners generally fall under the Next-best-view approaches as in~\cite{pito1999solution}\cite{bircher2016receding} \cite{dang2020autonomous} and have seen great application success, but other methods in similar directions exists, such as ERRT~\cite{lindqvist2021exploration} takes into consideration also actuation effort along with information gain in order to yield more efficiency towards exploration of unknown and unstructured areas. Additionally, the Rapid exploration method proposed in~\cite{cieslewski2017rapid} is developed to maintain a high MAV velocity, while exploring. Autonomous inspection of structures by utilizing a frontier based algorithm, along with a Lazy Theta$^*$ path planner, is presented in~\cite{faria2019applying}. Finally, an information driven frontier exploration method for MAV, which uses a hybrid approach between control sampling and frontier based is presented in~\cite{dai2020fast}. As state-of-the-art exploration method presented in~\cite{dharmadhikari2020motion} is tailored and deployed in large-scale exploration mission both in simulations and real world experiments. The developed planner is structured around motion primitives that search for admissible paths, taking advantage of efficient volumetric mapping with collision checks and future-safe path search that evaluates the variation of speed along the path, while also maximizing the exploration gain for an overall fast navigation scheme. Moreover, in~\cite{selin2019efficient} an exploration approach that combined frontiers with receding horizon next-best-view planning has been proposed. The frontiers are part of the global planning part, while the next-best-view is responsible for the local exploration part. 
In~\cite{xu2021autonomous} a dynamic exploration planner (DEP) for MAV exploration, based on Probabilistic road map has been presented. The sampling nodes are added incrementally and distributed evenly in the explored region, while the planner uses Euclidean Signed Distance Function map to optimize and refine local paths. The exploration scheme in~\cite{brunel2021splatplanner} presented the Permutohedral Frontier Filtering, which is based on bilateral filtering with permutohedral lattices to extract the score-based spatial density of the selected frontiers. Multiple studies have also incorporated visual servoing based path planning and control architectures for mobile robots as presented in \cite{dirik2020visual}. The authors in \cite{donmez2018vision} have formulated gaussian functions based control architecture for mobile robot that rely on mainly visual information of surrounding. The authors have extended the work further in \cite{donmez2020design} that uses decision trees as well as adaptive potential area methods to achieve autonomous control of mobile robots in real life applications. In the field of sampling based space mapping area the research studies presented in \cite{donmez2017bi}, uses bi RRT method to smooth the RRT path using curve fitting methods. In \cite{donmez2017bi} the Ability to navigate from start to goal position using the smooth path by curve fitting also addresses the problem of actuation of robot if extended for MAV in future. More path planning and control approaches have been discussed in a case study presented by the authors in \cite{okumucs2020cloudware} discuss in detail the planning and control of automated ground vehicles in industry. 

Various planning algorithms have been developed for navigation of aerial platforms in unknown environments, where in general they can be divided in map-based or memory-less approaches or their combination. In~\cite{ryll2019efficient} a hierarchical planning framework that combines map building from fused depth data, as well as instantaneous depth data, both organized into separate K-D trees has been proposed. The planner relies on a slower global planner to get a goal location, which is evaluated using motion primitives against the K-D trees with the lowest cost candidate primitive to be selected.  In~\cite{zhang2019maximum} a motion planning method for fast navigation of autonomous MAVs has been developed. The algorithm divides the environment modeling in two parts: i) the deterministically visible area within the on board sensor range, and ii) the probabilistically known area beyond the sensor range from a-priory map. The planning method maximizes the likelihood of reaching a goal, where a finite set of candidate trajectories are separated into groups and evaluated for collisions. In~\cite{matthies2014stereo} a navigation method for MAVs based on disparity image processing has been proposed. More specifically, the disparity image is used for direct collision checking, incorporating C-space expansion of obstacles. The motion planning part verifies obstacle free trajectory, projecting them into the disparity image and comparing their disparity values with the C-space disparity values for collision checking. In~\cite{bucki2020rectangular} a memory-less planner that is partitioning free space in pyramids, using direct depth image measurements has been demonstrated. The use of spatial generation of pyramids of the free spaces, allows for labeling obstacle free trajectories that lie inside the pyramids, while achieving fast generation of large number of candidate trajectories and performs collision checks. In~\cite{ahmad20213d} the authors present a reactive navigation system for MAV exploration. The developed algorithm is based on a two layered planning architecture that leverages the global environment map for frontier generation and local instantaneous sensor data for obstacle avoidance based on artificial potential fields. In~\cite{tordesillas2021faster} “FASTER” has been developed, an optimization based planning approach for fast and safe motion in unknown environments. The planner leads to high-speed navigation by allowing to plan in known and unknown configuration space using a convex decomposition in a two trajectory design approach, a fast and a safe trajectory. In \cite{lindqvist2021reactive} a reactive navigation and collision avoidance scheme for MAVs that combines a layer of obstacle detection based on 2D LiDAR with NMPC constraints was proposed for agile local navigation. In~\cite{kanellakis2020look} a collection of sensor based heading regulation methods have been proposed for aerial platform navigation along underground tunnel areas. In this work the heading regulation methods  using i) image centroid calculation from either single image depth estimation, or dark area contour extraction, or CNN for dark area extraction and ii) from processing 2D lidar measurements have been described. In~\cite{florence2018nanomap} a mapping for motion planning architecture that queries for the minimum-uncertainty view of a point in space, searching a set of recent depth measurements under noisy relative pose transforms has been presented. This work enables the identification of local 3D obstacles in the presence of significant state estimation uncertainty, evaluating motion plans. Table~\ref{table:1} summarizes the SoA exploration strategies, while highlighting the contribution of REF.

\begin{table}[h!]
\small
\renewcommand{\arraystretch}{1.3}
\caption{\bf Different exploration frameworks and their corresponding exploration-planning approach}
\label{table:1}
\centering
\begin{tabular}{|c|c|} 
 \hline
 \bfseries Framework & \bfseries Exploration approach \\  
 \hline\hline
    \rowcolor[HTML]{E6E6E6} 
    \textbf{\cite{yamauchi1997frontier}} & \begin{tabular}[c]{@{}c@{}} Closest frontier based on euclidean distance and \\ navigation to selected frontier based on depth-first-search on grid \end{tabular} \\
    \rowcolor[HTML]{FFFFFF}
    \textbf{\cite{zhou2021fuel}} & \begin{tabular}[c]{@{}c@{}} Incremental frontier structure and hierarchical planning for \\ trajectory generation to selected frontier\end{tabular} \\
    \rowcolor[HTML]{E6E6E6} 
    \textbf{\cite{zhu20153d}}              & \begin{tabular}[c]{@{}c@{}} Maximize information gain based on travel cost to frontier \end{tabular} \\
    \rowcolor[HTML]{FFFFFF}
    \textbf{\cite{cieslewski2017rapid}}   & \begin{tabular}[c]{@{}c@{}} Selection of furthest frontier in FOV to maintain high speed flight \&\\ switches to classical frontier approach when no frontiers exist in FOV \end{tabular} \\
    \rowcolor[HTML]{E6E6E6}
    \textbf{\cite{williams2020online}} & \begin{tabular}[c]{@{}c@{}} Exploration derived from direct point cloud visibility \\ to reduce mapping computation \end{tabular} \\
    \rowcolor[HTML]{FFFFFF}
    \textbf{\cite{fraundorfer2012vision}}   & \begin{tabular}[c]{@{}c@{}} Compute for centroid of closest frontier cluster \&\\ polar histogram based computation for cost to reach selected frontier \end{tabular} \\
    \rowcolor[HTML]{E6E6E6}
    \textbf{\cite{lindqvist2021exploration}} & \begin{tabular}[c]{@{}c@{}} Sampling based RRT structure approach to \\ maximize information gain with minimizing actuation cost \end{tabular} \\
    \rowcolor[HTML]{FFFFFF}
    \textbf{\cite{dharmadhikari2020motion}}  & \begin{tabular}[c]{@{}c@{}} 3D acceleration sampling to compute collision free paths to \\ maximum volumetric gain vertices using motion primitives \end{tabular} \\
    \rowcolor[HTML]{E6E6E6}
    \textbf{\cite{dang2019graph}}  & \begin{tabular}[c]{@{}c@{}} Bifurcated Local and global exploration approach. \\ Sampling based graph for local exploration \&\\ global re-positioning to closest \end{tabular} \\
    \rowcolor[HTML]{FFFFFF}
    \textbf{\cite{reinhart2020learning}}  & \begin{tabular}[c]{@{}c@{}} Learning based exploration derived from \\ graph based planning exploration planning architecture. \end{tabular} \\
    \rowcolor[HTML]{E6E6E6}
    \begin{tabular}[c]{@{}c@{}}\textbf{[REF]} \end{tabular}  & \begin{tabular}[c]{@{}c@{}} Safe frontier generation for local and global exploration \&\\ local frontier selection based on heading and avoidance cost \&\\ heading regulation, height difference and travel to frontier cost \\ based global re-positioning when local exploration gain is low \end{tabular} \\
    
 \hline
\end{tabular}
\normalsize
\end{table}

\subsection{Contributions}
The exploration, global planning and navigation architecture of this work is part of the development efforts within the COSTAR team~\cite{agha2021nebula},\cite{nikolakopoulos2021pushing} related with the DARPA Sub-T competition~\cite{subtworld}, while it is directly applicable for cave environments. Based on the above mentioned state-of-the-art, the key contributions to this article are listed in the following manner.
\begin{itemize}
\item The main contribution of this work stems from the development of safe frontier points generation and local as well as global cost based candidate frontier point selection method. In the presented work we extend the classical and rapid frontier exploration approaches with the improvements concerning safety of MAVs in field as well as maintaining agile nature of exploration. The proposed approach focuses on local frontier selection that take into account the position of such frontier relative to any static or dynamic obstacle in the field of view while also minimizing yaw movement of MAV. When no such frontier exist in the local field of view, the global re-positioning of the MAV is triggered in order to lead the MAV to global frontiers that lead the MAV to more unexplored space. The global re-positioning method is formulated such that it associate cost based on the overall actuation effort required by the MAV to move to a global frontier. The proposed global re-positioning of the MAV considers various factors such as MAV safety, actuation cost as well as how much of the unexplored space will be seen from a potential global frontier. Such contribution differentiate our method from other rapid frontier exploration approaches that directly switch to classical frontier approach, instead in our method MAV globally re-positions itself based on multi layer cost assignment in global frontier selection. As it will be presented, such contribution is particularly important in multi branched tunneling or caving system exploration scenarios. 

\item The second contribution presented in this article is the development of overall autonomy framework which addresses the problem of exploration, safety margin based path planning and reactive navigation through nmpc based control of MAVs. The dedicated risk aware path planning and potential fields based avoidance scheme incorporated within the proposed framework allows to push the limits of exploration in the candidate goal selection process in wide, narrow and obstructed  environments. Extensive simulations are performed for validating the proposed framework in multiple difficult scenarios in order to benchmark the safety, speed and versatility the presented autonomy modules. 

\end{itemize}
The rest of the article is structured as follows.
\autoref{problem} presents the problem formulation considered in this work. \autoref{sec:proposed} presents the proposed safe frontier points generation as well as intelligent goal selection with the focus on safe yet fast autonomous exploration addressing the minimizing actuation effort of the MAV. The section also describes the overall autonomy framework which is the combination of exploration, global path planning as well as nmpc based reactive navigation.  In \autoref{sec:sim_results}, detailed analysis on simulation experiments are presented that prove the efficacy of the proposed scheme. Finally \autoref{sec:conclusions} provides a discussion with concluding remarks of the proposed work.

\section{Problem Formulation}\label{problem}

The problem considered in this work is exploration of bounded  3D space denoted as $V \subset \mathbb{R}^3$. Occupancy probability based modelling is adapted in order to model free, occupied and unknown space around the robot. The provided framework is targeted for a direct use case in field robotics where autonomous robots are deployed in cave and mines for rapidly mapping the areas. In the case of exploration of bounded space, the exploration will be considered complete when $V_{occupied} \bigcup V_{free} = V$ while $V_{unknown} = \varnothing$. In the proposed approach, exploration process is subject to vehicle actuation effort, limitations in time of flight as well as risk margin for safe path generation. In order to be deployed in real scenario, the exploration and planning framework should adapt based on the available computational resources. The performance evaluation of the proposed framework will be based on exploration time as well as actuation efforts of the MAV.


\section{Proposed Approach}\label{sec:proposed}
%
The proposed approach employs a frontier based exploration technique which is modified with the focus on making exploration fast, safe and versatile based on available computational resources. We use occupancy grid maps as a mapping framework, which can generate a 2D or 3D probabilistic map. A value of occupancy probability is assigned to each cell that represents a cell to be either free or occupied in the grid. In this work we are targeting 3D exploration of bounded and unbounded space therefore using the baseline framework of OctoMap~\cite{hornung2013octomap} we build on top of it in order to develop the proposed 3D occupancy checking framework used in this work. Let us denote a voxel as $v(x,y,z)$. The voxel $v$ is subdivided into eight smaller voxels until a minimum volume is reached. The minimum volume is same as the octree resolution $v_{res}$. Corresponding to each sensor update if a certain volume in the octree is measured and if it is observed to be occupied, the node containing that particular voxel is marked as occupied. Using ray casting operation for the nodes between the occupied node and the origin (sensor), in the line of ray, can be initialized and marked as free. This process leaves the uninitialized nodes to be marked unknown until the next update in the octree. Let us denote the estimated value of the probability $P(N\ |\ z_{1:t})$ of the node $N$ to be occupied for the sensor measurement $z_{1:t}$ by:

\begin{align*}
    P(N | z_{1:t}) = [1 + \frac{1-P(N|z_{t})}{P(N|z_{t})} \frac{1-P(N|z_{1:t-1})}{P(N|z_{1:t-1})} \frac{P(n)}{1-P(n)}]^{-1}
\end{align*}

where $P_{n}$ is the prior probability of node $N$ to be occupied. 
Let us denote the occupancy probability as for node N to be occupied as $P_{N}^{o}$
$$
v(x,y,z)=\begin{cases}
                    Free, & \text{if $P^{o} < P_{n}$}\\
                    Occupied, & \text{if $P^{o} > P_{n}$}
                \end{cases}
$$                

\begin{table}[h!]
\small
\renewcommand{\arraystretch}{1.3}
\caption{\bf Description of the notations used in proposed methodology}
\label{table:notations}
\centering
\begin{tabular}{|c|c|} 
 \hline
 \bfseries Notation & \bfseries Meaning \\  
 \hline\hline
    $\mathcal{\{F\}}$ & All frontier set \\ 
    $\mathcal{\{O\}}$ & Occupied nodes set \\
    $\mathcal{\{C\}}$ & Valid frontiers set \\ 
    $\mathcal{\{SF\}}$ & Safe frontiers set \\
    $\mathcal{\{L\}}$ & Local frontiers set \\
    $\mathcal{\{G\}}$ & Global frontiers set \\
    $R$ & Sensor measurement range \\
    $r$ & cleaning radius \\
    $m$ & risk margin \\
    $v_{res}$ & octree resolution \\
    $V_\beta$ & Horizontal FOV \\
    $V_\beta$ & Vertical FOV \\
    $f_{(x,y,z)}$ & Frontier position \\
    $C_{(x,y,z)}$ & MAV current position \\

 \hline
\end{tabular}
\normalsize
\end{table}

Let us define the sensor range $R$ and a sphere of radius $r$ around the MAV. This radius $r$ will be denoted as cleaning radius from here after. The after each update in current octree, if a frontier lie inside this sphere, the frontier is marked as seen and such frontier is deleted from $\mathcal{\{F\}}$. The cleaning radius is defined such that $r\ <\ R$ therefore new frontiers will always be generated at distance $R$ and as the MAV navigate towards frontier, the frontiers lying within the sphere of radius $r$ are deleted and less number of frontiers need to be iterated through in candidate goal selection process. The iterator is defined as $it$. The meanings of the important notations used in this work is presented in \autoref{table:notations}. 

\begin{algorithm}
\SetAlgoLined
\caption{Safe Frontier Generation} \label{generation}
\KwInput {$v_{res}, Current\ octree, k, r$}
\KwOutput {$\{\mathcal{F}\}$,$\{\mathcal{O}\}$}
\For{$N : Current\ octree$}{
    \eIf{$P^{o}_{N} < P_{n}$}{
        \If{$N.distance() < r$}{
            $it \gets 0;$ \\
            \For{$Neighbours : N.getNeighbour()$}{
                \eIf{$P^{o}_{N} > P_{n}$}{
                $it \gets 0;$ \\
                $break;$}
                    {
                    $it \gets it + 1$
                    }
            }
        }
        \If{$it \geq k$}{
             $\{\mathcal{F}\}.add(it\_N);$
        }
    }
    {$\{\mathcal{O}\}.add(it\_N);$}
    
        \For{$N : \{\mathcal{F}\}$} {
            \If{$(N_{adj} + N_{m*v_{res}}) \notin \{\mathcal{O}\}$}{$\{\mathcal{SF}\}.add(N);$} 
        }
}
\end{algorithm}

\begin{figure}[h!]
  \centering
  \captionsetup{justification=centering}
    \includegraphics[width=\linewidth]{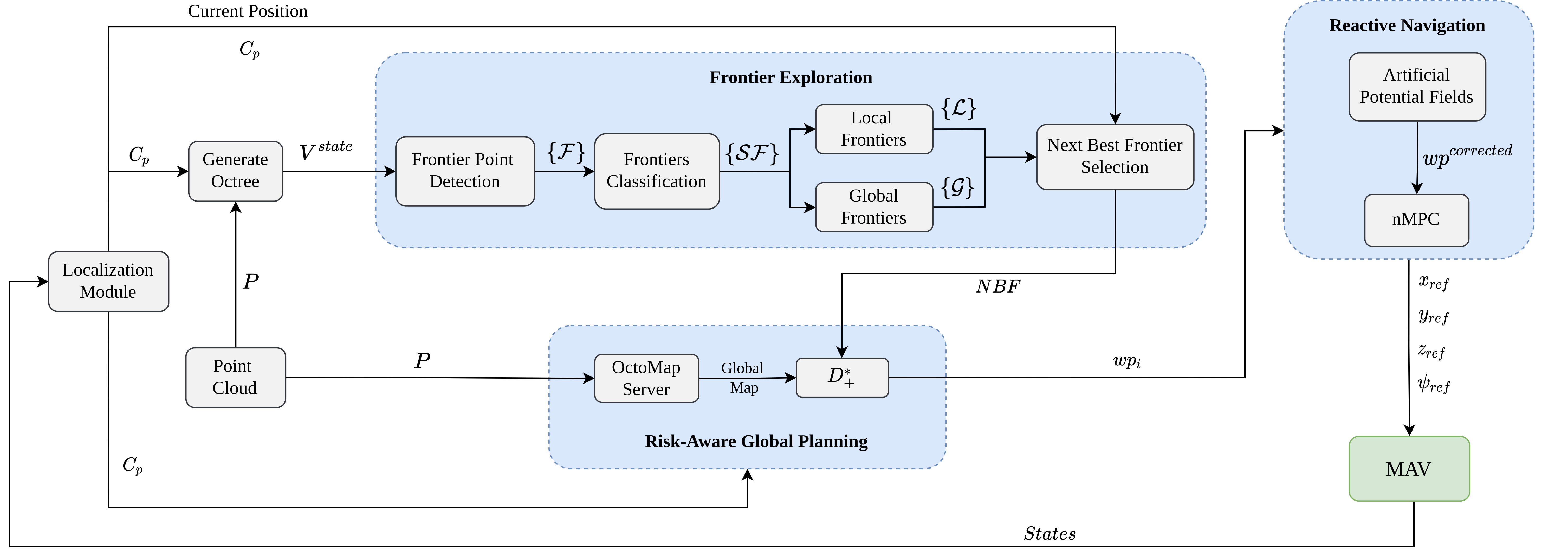}
    \caption{The proposed overall autonomy and navigation scheme}
  \label{fig:navigationframework}
\end{figure}

The exploration framework presented in this work is is made up of three essential modules, namely the safe frontier point generator, the cost based frontier point selection incorporating also collision check and finally candidate goal publisher as presented in \autoref{fig:navigationframework}. 

The first module takes the Lidar point cloud as input and based on the occupancy probability formulation as mentioned earlier, converts the sensor measurement in order to form an octree. The octree is defined as tree data structure in which each sub node is further divided into eight quadrants until minimum volume is reached. The safe frontier point generator module generates all safe frontiers based on the octree as depicted in algorithm \autoref{generation}. Let us define a risk margin parameter $m$ related to the voxel grid resolution $v_{res}$. At any instance in the exploration if node $N$  being currently checked for to be considered as a safe frontier then we also check the neighbouring adjacent nodes defined as $N_{adj}$ within the safety margin $m$. 

In our approach we formulate an additional layer of requirement in which we check the neighbouring Voxels of an uninitialized (Unknown) Node $N$ as described earlier and $\forall(N_{adj} + m*v_{res})$ if $(P^{o}_{N_{adj}} \leq P_{n})$ than the Node $N$ is considered as safe frontier node and is added to $\mathcal{\{SF\}}$, where $\mathcal{\{SF\}}$ is a set containing all safe frontiers. This means that a particular node $N$, it's adjacent node $N_{adj}$ as well as all nodes in the neighbourhood of node $N$ within the range of $m*v_{res}$ are checked and if all such nodes are seen to be free than the node $N$ is considered to be a safe frontier. To be marked as a frontier, each node should have at least $n$ number of minimum unknown or free adjacent nodes. This process makes a big difference in the computational complexity of the process because by specifying a certain risk margin $m$ and minimum unknown or free neighbours $k$ at the start of exploration, the trade off can be made between number of iterations and coverage quality. Another improvement our approach presents is that by not allowing any frontier to be close enough to an occupied node in context of risk margin, we guarantee that inaccessible frontiers can be eliminated which are generated due to the error in probabilistic occupancy mapping. The inaccessible frontiers are defined as the frontiers that are not safe to reach or impossible to reach in terms of MAV size and dynamics to pass through small openings in the map. This simply implies that risk margin can be set in correspondence with the size of the MAV such that the inaccessible areas can be patched and modelled as occupied in the map. The parameters $m$ as well as $k$ are proposed with the focus of testing the proposed approach in extremely difficult areas such as caves and mines where safety of the robot is a major concern.

\begin{algorithm}
\SetAlgoLined
\caption{Frontier Classification Based on Local or Global Exploration} \label{selection}
\KwInput  {$\{\mathcal{SF}\}$ $k$, $r$, $\alpha$ $\theta$ }
\KwOutput {$NBF$, $\{\mathcal{L}\}$, $\{\mathcal{G}\}$} 
\For{$N : \{\mathcal{SF}\}$}{
    \If{$N.distance() < r$}{
         $it \gets 0;$ \\
        \For{$Neighbours : N.getNeighbour()$}{
            \eIf{$Neighbour.isOccupied()$}{
             $it \gets 0;$ \\
             $break;$}
                {
                 $it \gets it + 1;$
                }
        }
        \eIf{$it < k$}{
             $\{\mathcal{SF}\}.remove(N);$
        }
            {
                 $\{\mathcal{C}\}.add(it\_N)$
            }
                \For{$ N : \{\mathcal{C}\}$}{
                \eIf{$(\alpha < (H_{\theta})/2)\ \& \ (\gamma < V_{\beta})$}{
                     $\{\mathcal{L}\}.add(it\_N)$}
                    {
                         $\{\mathcal{G}\}.add(it\_N)$}
                \If{$\{\mathcal{L}\} \neq \varnothing$}{
                    \For{$N : \{\mathcal{L}\}$}
                    {$NBF \gets  \argmin\limits_{\{N \in \{\mathcal{L}\}\}} (\mathcal{E})_{local}$}
                }
                \If{$\{\mathcal{L}\} = \varnothing, \{\mathcal{G}\} \neq \varnothing$}{
                    \For{$N : \{\mathcal{G}\}$}{
                     $NBF \gets  \argmin\limits_{\{N \in \{\mathcal{G}\}\}} (\mathcal{E})_{global}$}
                }
                \If{$\{\mathcal{L}\} \bigcup \{\mathcal{G}\} = \varnothing$}{
                     $it \gets 0;$ \\
                     $break;$ \\
                     $D^{*}_{+}.ComputeHomingPath()$ \\
                     $nMPC \gets HomingPath$
            }
        }
    }
}
\end{algorithm}

\begin{figure}[h!]
  \centering
  \captionsetup{justification=centering}
    \includegraphics[width=0.85\linewidth]{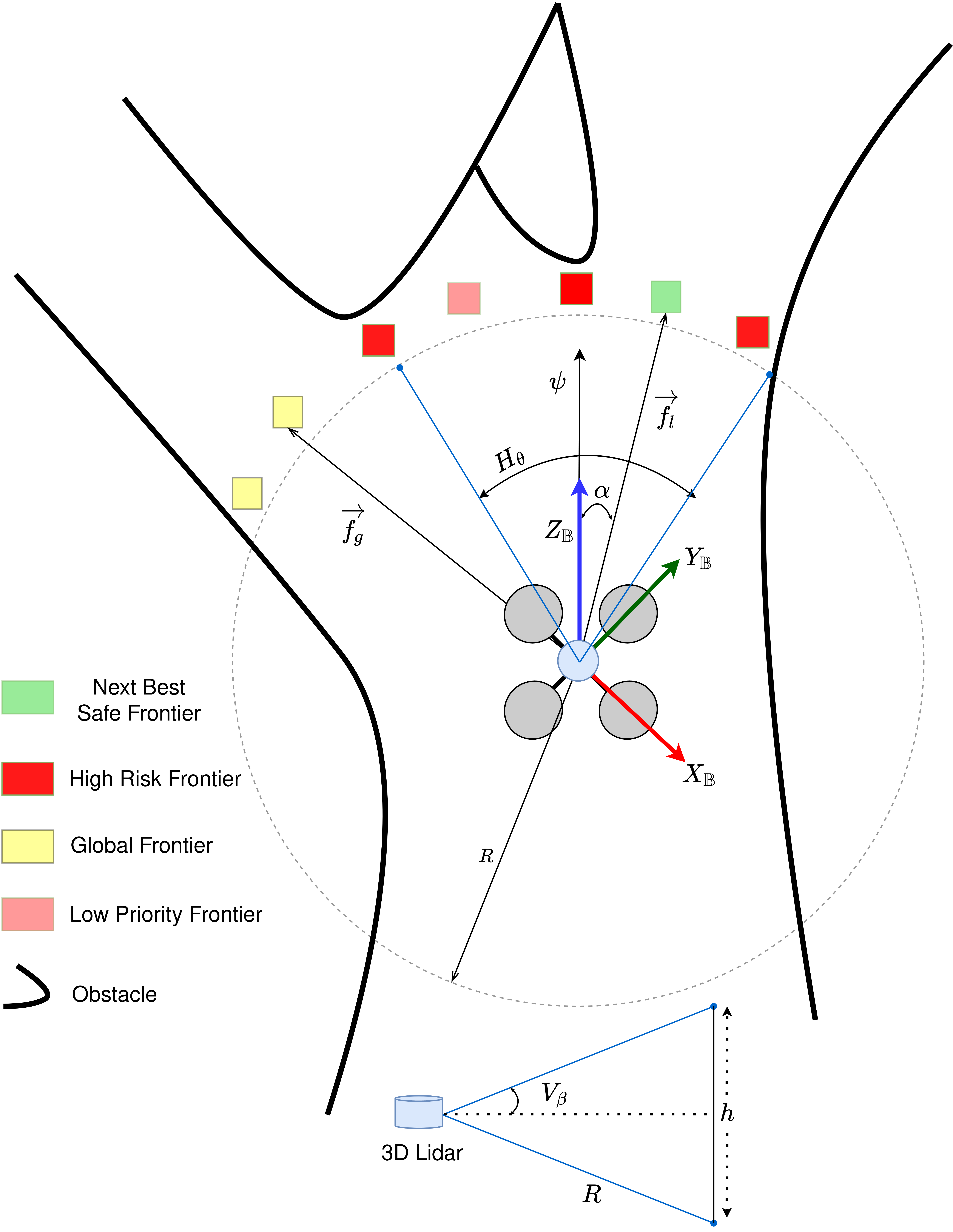}
    \caption{Frontier classification and notations used in the proposed framework}
  \label{fig:localglobal}
\end{figure}

As defined in \autoref{selection}, corresponding to each new sensor measurement we check if a $N \in \{\mathcal{F}\}$ is still a frontier. We define a candidate frontier set denoted as $\{\mathcal{C}\} \subset \{\mathcal{F}\}$ which contains all the valid frontiers which will be examined based on the MAV's position. A 3D Lidar is used in the proposed method to get sensor point cloud and thus, the framework generates frontiers in all directions surrounding the MAV but limited in the vertical directions with field of view $V_\beta$. In \autoref{selection}, we classify the frontier nodes in two further sets $\{\mathcal{L}\}, \{\mathcal{G}\} \bigcup \{\mathcal{C}\}$ named as Local and Global set respectively. Such Local and Global sets contain frontier nodes classified based on the selected horizontal and vertical field of view $H_\theta$ and $V_\beta$ respectively as shown in \autoref{fig:localglobal}.

This process allows us to prioritize the unknown space lying ahead of the MAV and if there exist no unknown space ahead of the MAV, the candidate goal is selected based on the global cost based goal assignment. 

$ \forall \ f \in \{\mathcal{L}\}$ are computed for extracting $NBF$ such that $\alpha \in [-\pi,\pi]$ is minimum. The frontier points from occupancy formulations are generated in the world frame ($\mathbb{W}$) but the frontier vector $\Vec{f}$ is calculated relative to the MAV body frame $\mathbb{\{B\}}$. As shown in \autoref{fig:localglobal}, the angle $\alpha$ is calculated with respect to $\mathbb{B}$. If a frontier $f$ and MAV's current position in inertial frame $\mathbb{W}$ is defined as $f(x,y,z)$ and $C(x,y,z)$ respectively then the angle $\alpha$ and $\gamma$ with respect to body frame $\mathbb{B}$ can be computed as,

\begin{equation}
    \alpha = tan^{-1}(\frac{f_{y} - C_{y}}{f_{x} - C_{x}}) - \psi
\end{equation}

\begin{equation}
    \gamma = cos^{-1}(\frac{h}{2*(f_{z} - C_{z})})
\end{equation}

where, $\psi$ is the heading angle of the MAV and $h$ is the vertical height of the footprint of 3D LiDAR field of view.

As discussed previously the \autoref{generation} also outputs a list of occupied nodes $\{\mathcal{O}\}$ which has occupancy probability $P^{o}$ higher than 0.5 thus considering the the cluster of occupied points lying in the field of view, the the frontier nodes having a lesser avoidance cost are also favoured to be the $Next Best Frontier$. The cost formulation for selecting a local or global candidate goal is as follows. If we define the current position of the MAV as $C(x,y,z)$ then the costs for local and global frontier selection can be formulated as,

\begin{equation}
    (\mathcal{\zeta})_{local} = 
    \overbrace{\frac{1}{\textbf{W}_{o} \sqrt{(p^{f}_{x} - p^{obs}_{x})^2 + (p^{f}_{y} - p^{obs}_{y})^2 + (p^{f}_{z} - p^{obs}_{z})^2} }}^\text{Avoidance cost} + \\  \overbrace{\textbf{W}_{h} * \alpha}^\text{Heading cost}
    \label{eqn:cost}
\end{equation}

\begin{align}
    (\mathcal{\zeta})_{global} = 
      \overbrace{\textbf{W}_{h} * \alpha}^\text{Heading cost} 
    + \overbrace{{\textbf{W}_{z} * ({f}_{z} - C_{z})}}^\text{Height difference cost} + \nonumber \\
    \overbrace{\textbf{W}_{d} \sqrt{({f}_{x} - C_{x})^2 + ({f}_{y} - C_{y})^2 + ({f}_{z} - C_{z})^2}}^\text{Distance cost}
    \label{eqn:costg}
\end{align}

Where, $\textbf{W}_{o}, \textbf{W}_{h}, \textbf{W}_{z}$ and $\textbf{W}_{d}  \in \mathbb{R}$ are defined as weights associated to avoidance, heading, height difference and distance cost respectively. We define the actuation effort $\mathcal{E}$ as a function of cost such that 
\begin{align*}
    \mathcal{E} = f(\mathcal{\zeta} + T_{hover})
\end{align*}
where, $T_{hover}$ is the minimum thrust required for hovering with zero torques about the MAV arms. Thus, by optimally selecting the next pose reference command for the MAV the actuation effort can be minimized. The MAVs consume high energy to produce yaw torque due to the motor saturation constraints while also keeping the MAV hovering.

The overall autonomy scheme of the proposed work is presented in \autoref{fig:navigationframework}. As discussed earlier, the framework uses 3D LiDAR or a camera depth cloud as point cloud input and upon point cloud filtering, the framework generates an octree of occupied, free and unknown nodes. Using the workflow described in \autoref{generation}, the framework detects frontier points and classifies set of safe frontiers. As presented in the autonomy and navigation scheme (\autoref{fig:navigationframework}), based on local or global frontier, the risk aware global planning module plans a collision free path to the next best frontier. The $NBF$ is then fed into the reactive navigation and control framework to actuate the MAV to navigate to selected frontier point. In \autoref{fig:navigationframework} APF stands for Artificial Potential Fields that we have incorporated with Nonlinear Model Predictive Control for collision avoidance. The baseline framework for reactive navigation and control used in this framework is inspired from our previous work~\cite{lindqvist2021compra}. The Next Best Frontier is sent to a risk aware global planning module which is the extension of $D^{*} Lite$ algorithm but implemented with octomap framework in this case. The global planning module $D^{*}_{+}$ uses the modelled occupied space in order to plan a safe path to the $NBF$. The risk margin formulation in an expandable octomap grid for global planning is presented in detail in our previous work~\cite{karlsson2021d}. 

\section{Exploration Mission Experiments}\label{sec:sim_results}

In order to validate and test the performance of our proposed exploration approach we use the M100 MAV provided in the open source Rotors Simulator~\cite{Furrer2016} framework. Next-Best-view~\cite{bircher2016receding} has been widely used for bench-marking the exploration-planning algorithms. In this work we compare our framework with the latest version of NBV, State-of-the-Art Motion Primitive Based planner (mbplanner)~\cite{dharmadhikari2020motion} which is developed also as part of the development efforts within DARPA Sub-T challenge. We use a custom cave model with multiple junctions, obstructed walls, narrow openings, steep slopes as well as tunnels with dead-ends for simulation. The cave environment has been made open sourced for public \cite{Patel2021}. For fair comparison all simulations are performed with same computational unit having Intel core i7 processor and 16 GB memory on ROS Melodic running on Ubuntu 18.04. For mbplanner also the simulations are performed using the cave virtual world where the tuning of parameters such as MAV velocity, mapping resolution and sampling time, was similar to the ones used for the proposed method.

\begin{figure}[h!]
  \centering
  \captionsetup{justification=centering}
    \includegraphics[width=\linewidth]{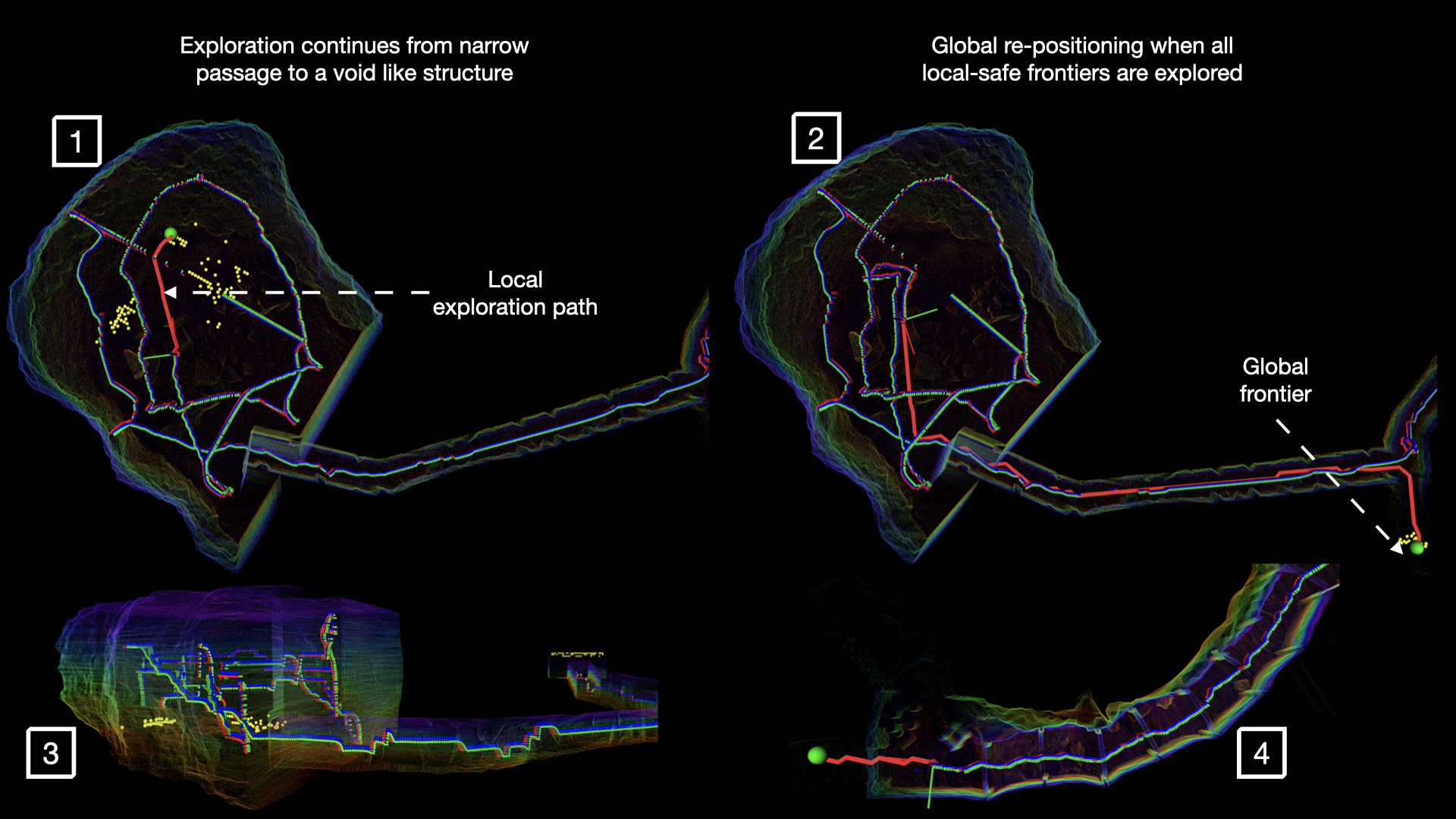}
    \caption{DARPA-Sub-T virtual world: Exploration of narrow-confined passages as well as large cave like voids using proposed framework. In (1,2,3) the rapid exploration-coverage nature of the proposed framework is shown. In (4) the safe way point selection and risk aware planning to safe frontier is shown}
  \label{fig:void}
\end{figure}

\begin{figure}[h!]
  \centering
  \captionsetup{justification=centering}
    \includegraphics[width=\linewidth]{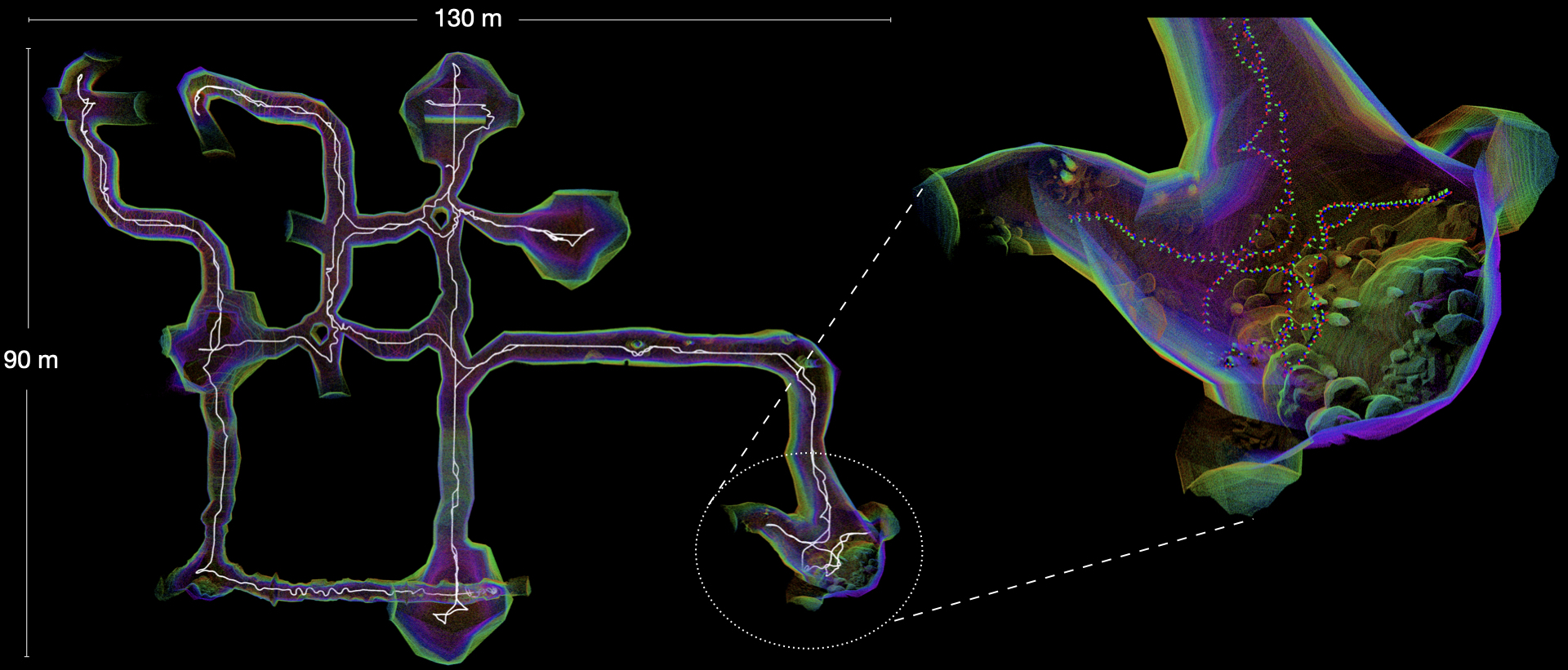}
    \caption{Multi-branched large 3D virtual cave world exploration using proposed framework}
  \label{fig:caveworld}
\end{figure}

In \autoref{fig:void} different exploration instances are shown. As described in \autoref{sec:proposed} the proposed framework (REF) also uses frontier cleaning radius and due to which coverage of large cave like voids can also be performed while exploring. Using the proposed framework the MAV is also able to navigate in narrow and obstructed passages and at the end of such passages if a void like area can also be covered efficiently. The simulation experiment is also carried out to explore a multi-branched virtual cave environment having narrow passages continuing in different heights for a true 3D exploration. The environment is also made open source \cite{akoval2020}. In \autoref{fig:caveworld} the exploration of the virtual cave environment is shown. 

\begin{figure}[h!]
  \centering
  \captionsetup{justification=centering}
    \includegraphics[width=\linewidth]{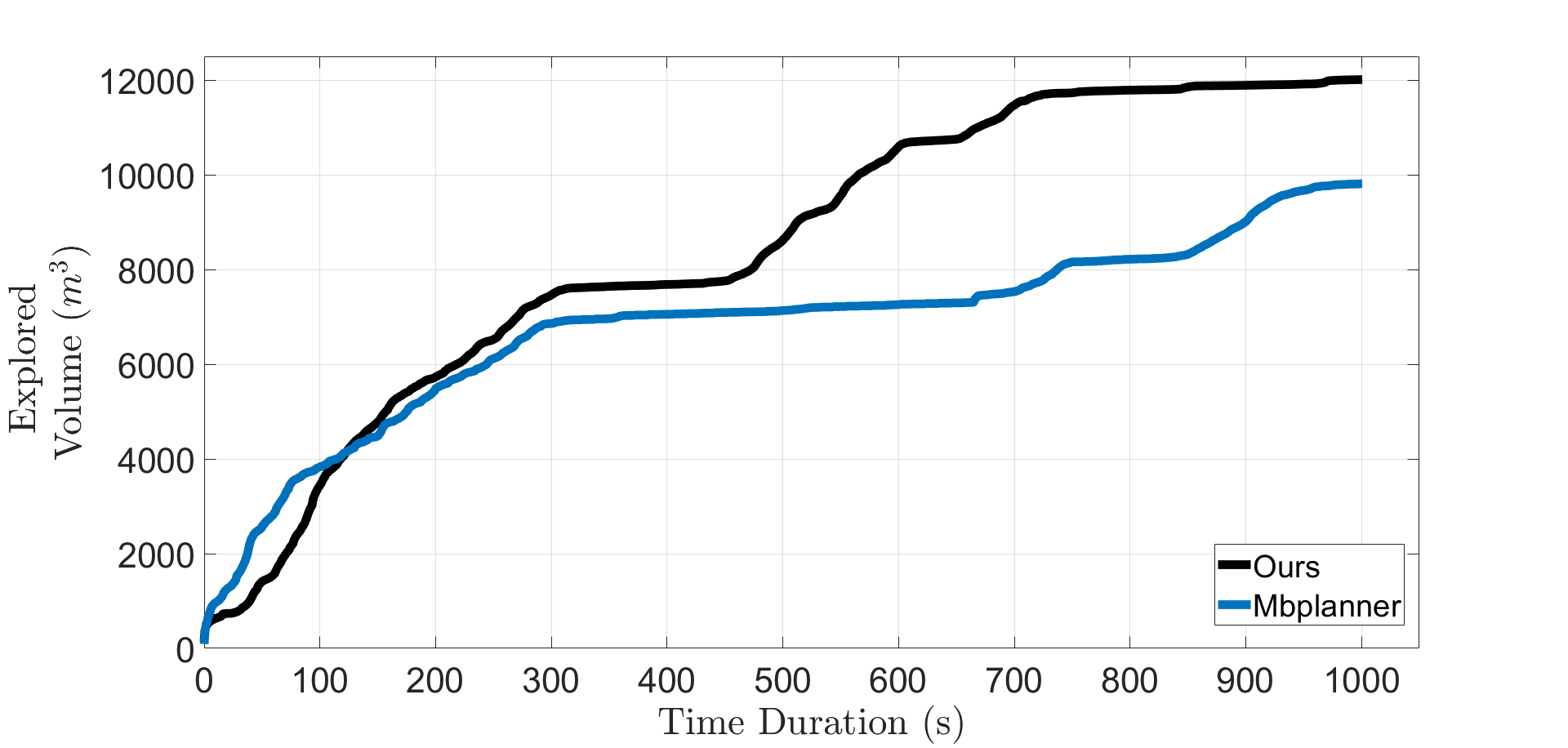}
    \caption{Volumetric gain by the two methods}
  \label{fig:volume}
\end{figure}

\begin{figure}[h!]
  \centering
  \captionsetup{justification=centering}
    \includegraphics[width=\linewidth]{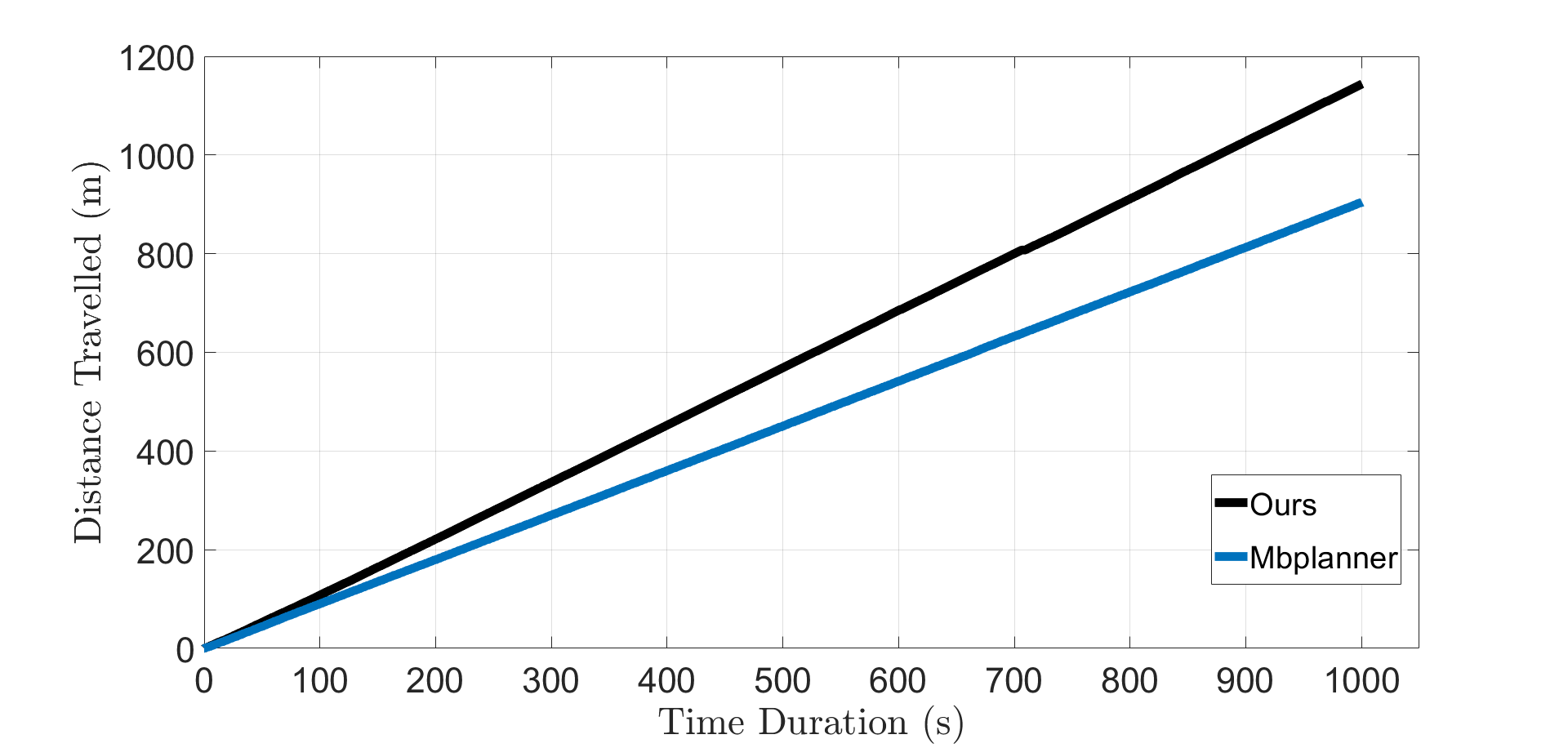}
    \caption{Distance covered by the two methods}
  \label{fig:distance}
\end{figure}

\begin{figure*}[h!]
    \centering
    \subfigure[]
    {
        \includegraphics[width=0.30\linewidth]{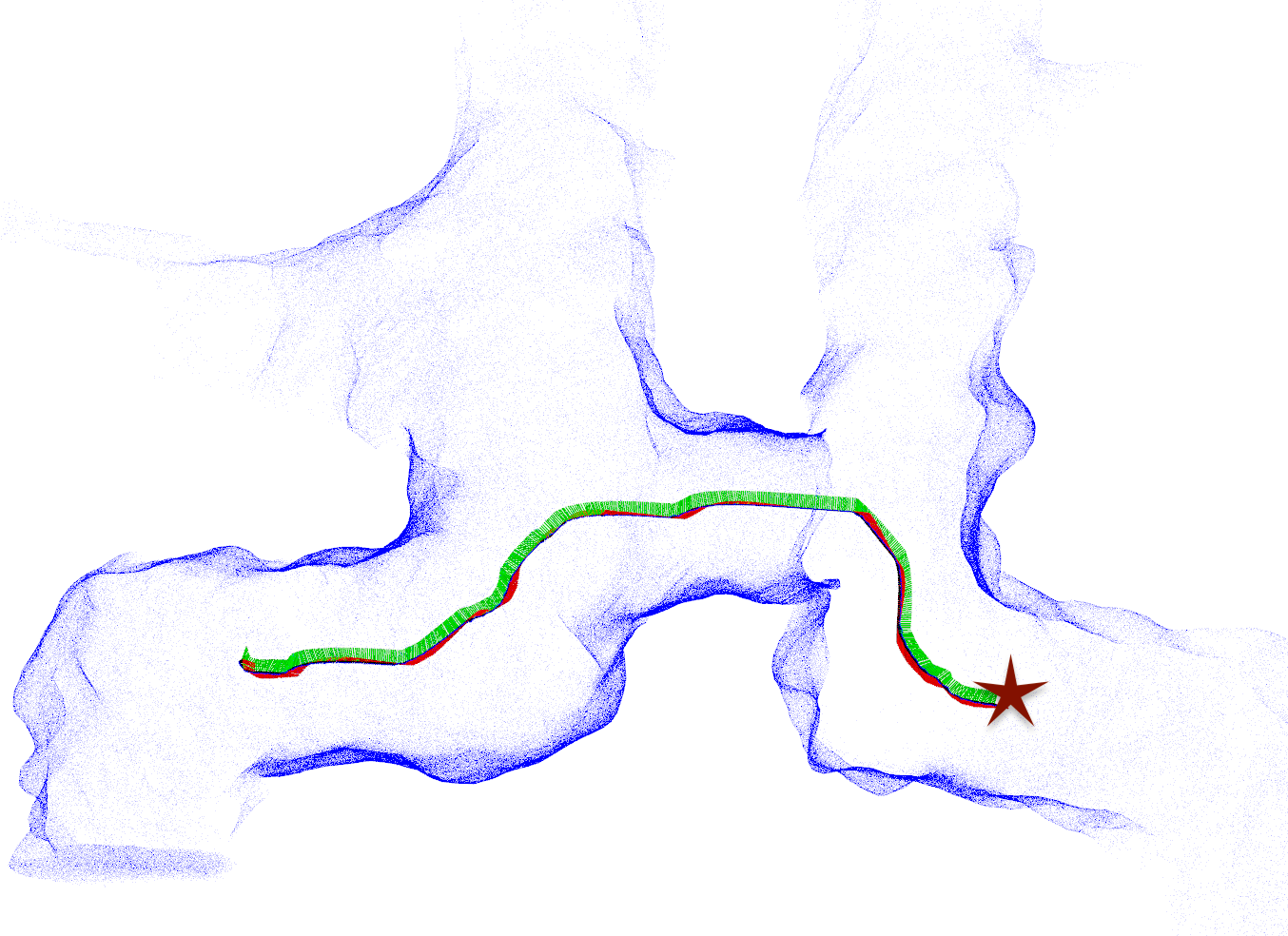}
        \label{fig:Ours : 1min}
        
    }
    \subfigure[]
    {
        \includegraphics[width=0.30\linewidth]{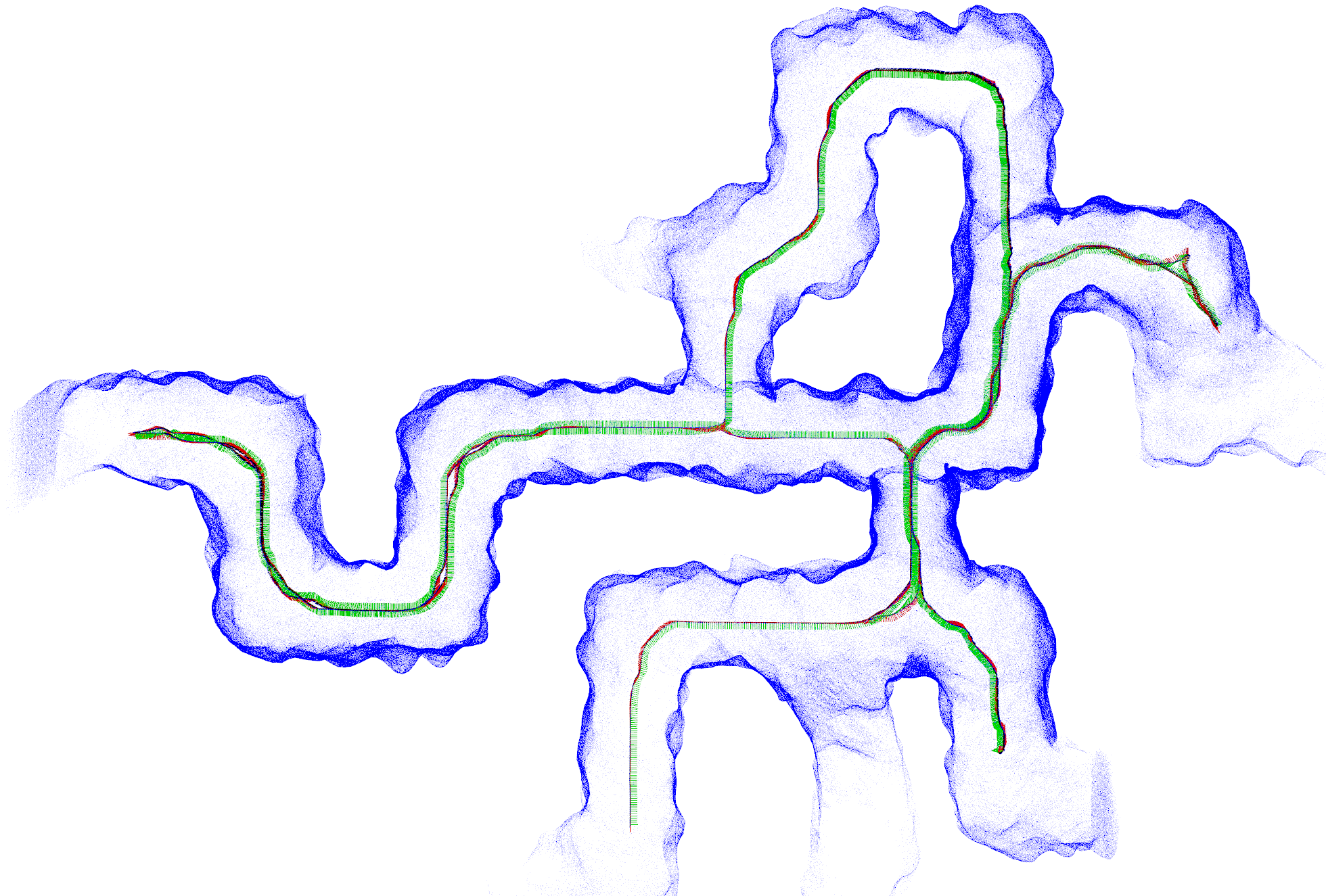}
        \label{fig:Ours : 5min}
        
    }
    \subfigure[]
    {
        \includegraphics[width=0.30\linewidth]{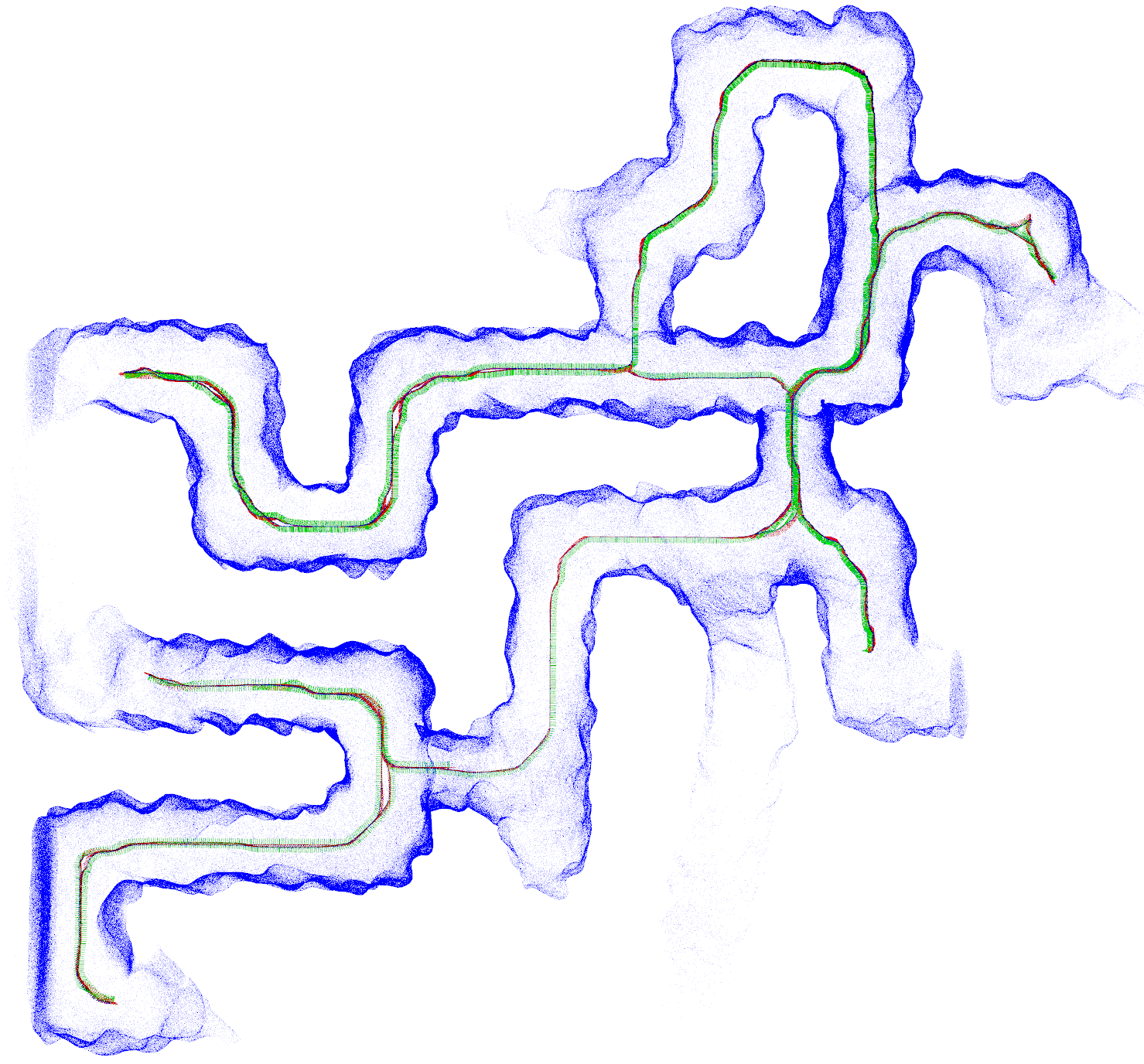}
        \label{fig:Ours : 10min}
        
    } 
    \\
    \subfigure[]
    {
        \includegraphics[width=0.30\linewidth]{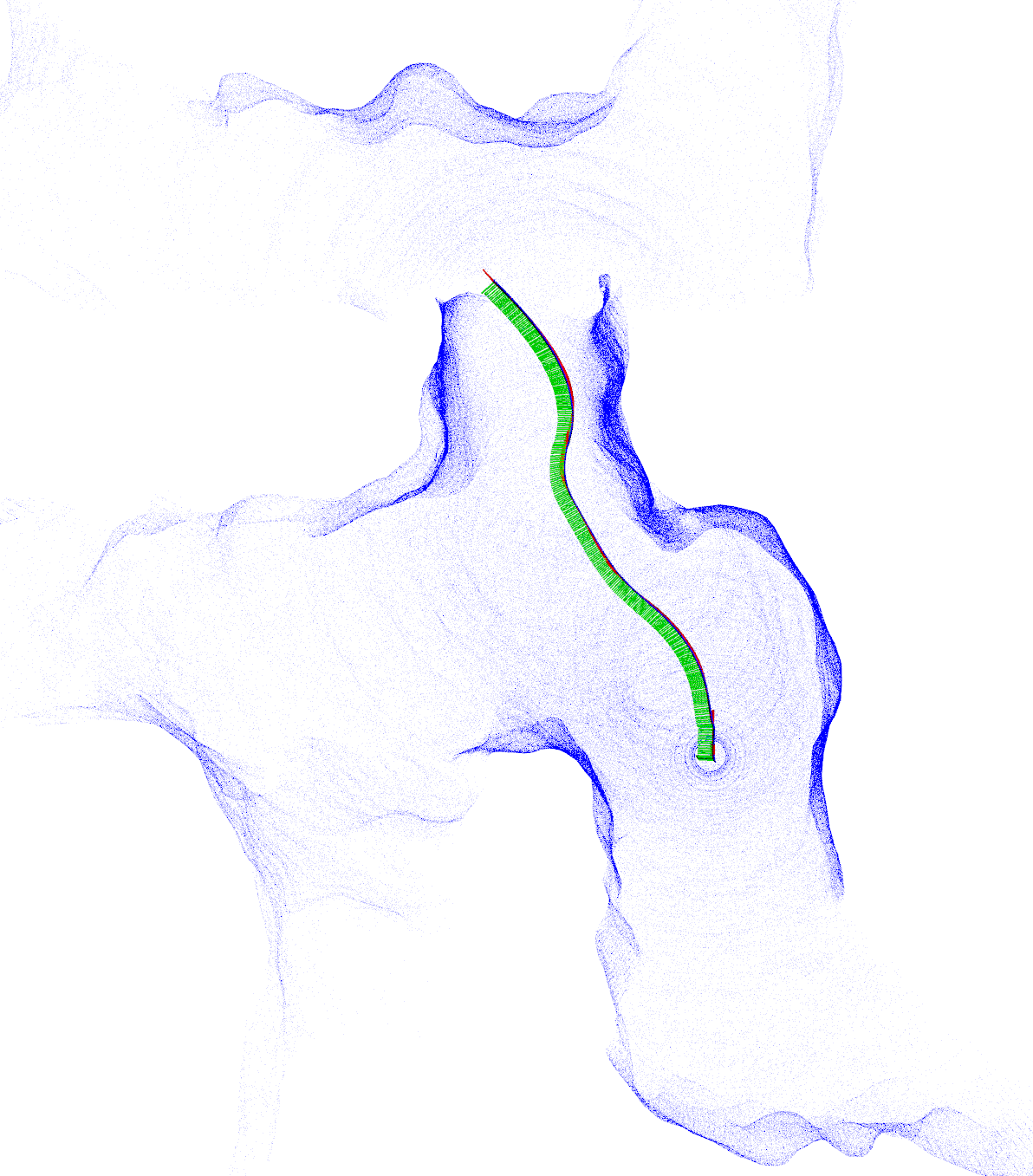}
        \label{fig:mbplanner : 1min}
        
    }
        \subfigure[]
    {
        \includegraphics[width=0.30\linewidth]{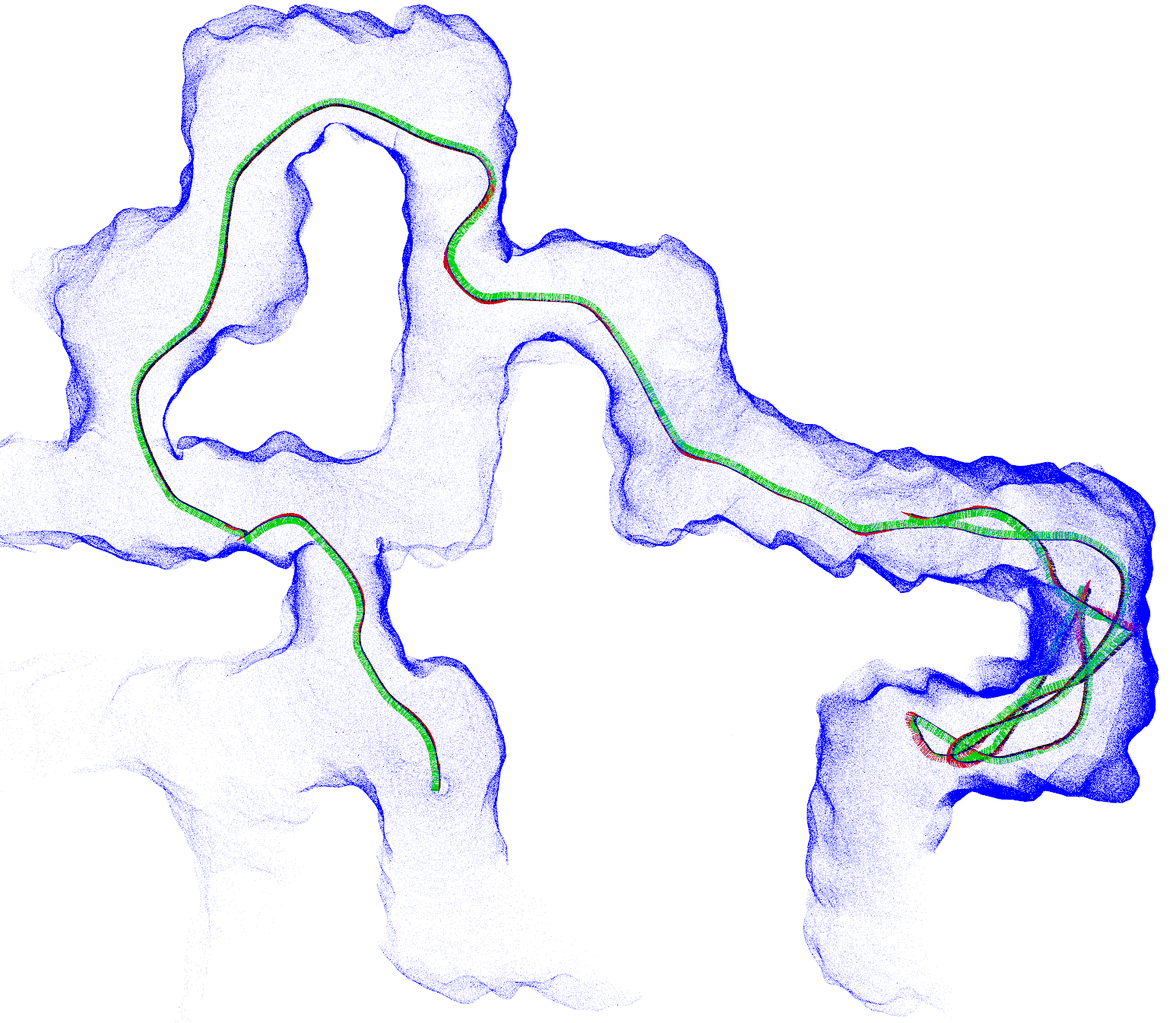}
        \label{fig:mbplanner : 10min}
        
    }
        \subfigure[]
    {
        \includegraphics[width=0.30\linewidth]{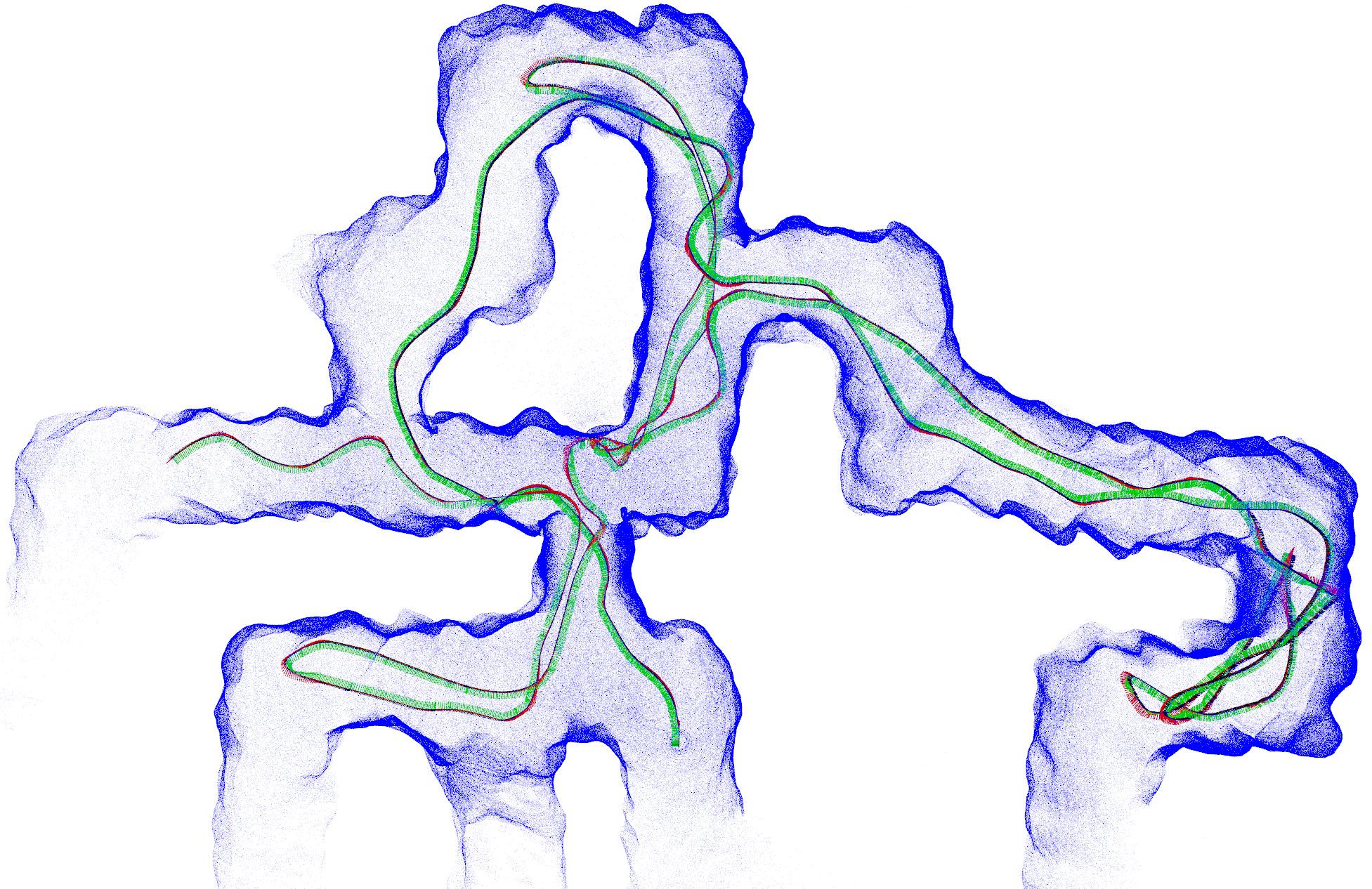}
        \label{fig:mbplanner : 20min}
    }
    \caption{Time based exploration : MAV trajectory (a) ours : 1 minute, (b) ours : 10 minute, (c) ours : 15 minutes, (d) mbplanner : 1 minute, (e) mbplanner : 10 minutes, (f) mbplanner : 20 minutes}
    \label{fig:explorationtrajectory}
\end{figure*}

In \autoref{fig:volume} and \autoref{fig:distance} the explored volume and distance covered by the two exploration frameworks is presented. \autoref{fig:volume} and~\autoref{fig:distance} depict that our method performs significantly close to the State-of-the-Art mbplanner in terms of exploration volume of the cave environment and distance covered respectively. The proposed approach achieves slightly higher explored volume for the same mission time, this is because of the novel Next Best Frontier selection approach as adapted in \autoref{sec:proposed}. As presented in \autoref{fig:explorationtrajectory}, MAV trajectory in our approach is significantly inline with the goal of maximizing the movement into unknown areas while limiting repeated visits to already mapped areas. In \autoref{table:comparison} the exploration volume and distance travelled by the MAV in multiple different runs with different mission duration is presented for both planning framework. As it is evident from \autoref{table:comparison} that the proposed Rapid Exploration Framework (REF) shows higher exploration volume as well as ground covered by the MAV in multiple different runs because of the nature of computing next paths while navigating to current path. All missions considered in \autoref{table:comparison} have same start positions for both planning framework and the MAVs do not return to base in considered cases therefore, showing the exploration capability comparison in given time with same configuration. 

\begin{table}[h!]
\small
\renewcommand{\arraystretch}{1.3}
\caption{\bf Exploration volume and distance from multiple runs}
\label{table:comparison}
\centering
\begin{tabular}{|c|c|c|c|c|} 
 \hline
 \bfseries \bfseries \begin{tabular}[c]{@{}c@{}} Mission \\ Duration \end{tabular} & \bfseries \begin{tabular}[c]{@{}c@{}} Volume \\ (REF) \end{tabular} & \bfseries \begin{tabular}[c]{@{}c@{}} Volume \\ (Mbplanner) \end{tabular} & \bfseries \begin{tabular}[c]{@{}c@{}} Distance \\ (REF) \end{tabular} & \bfseries \begin{tabular}[c]{@{}c@{}} Distance \\ (Mbplanner) \end{tabular} \\  
 \hline\hline
 100 s & 3578 $m^{3}$ & \textbf{3840 $m^{3}$} & \textbf{163 $m$} & 154 $m$ \\ \hline
 300 s & \textbf{7854 $m^{3}$} & 6958 $m^{3}$ & \textbf{284 $m$} & 236 $m$ \\ \hline
 600 s & \textbf{11367 $m^{3}$} & 8438 $m^{3}$ & \textbf{670 $m$} & 476 $m$ \\ \hline
 900 s & \textbf{14477 $m^{3}$} & 9851 $m^{3}$ & \textbf{1066 $m$} & 781 $m$ \\ \hline
 1200 s & \textbf{17524 $m^{3}$} & 12760 $m^{3}$ & \textbf{1185 $m$} & 962 $m$ \\ 
    
 \hline
\end{tabular}
\normalsize
\end{table}

However, it is also important to mention that even though the $V_{unknown}$ sampling approach in both method is different, the next way points in both case are selected with the focus on maximizing the information gain and exploration volume in same time. 

\begin{figure*}[h!]
    \centering
    \subfigure[]
    {
        \includegraphics[width=0.45\linewidth]{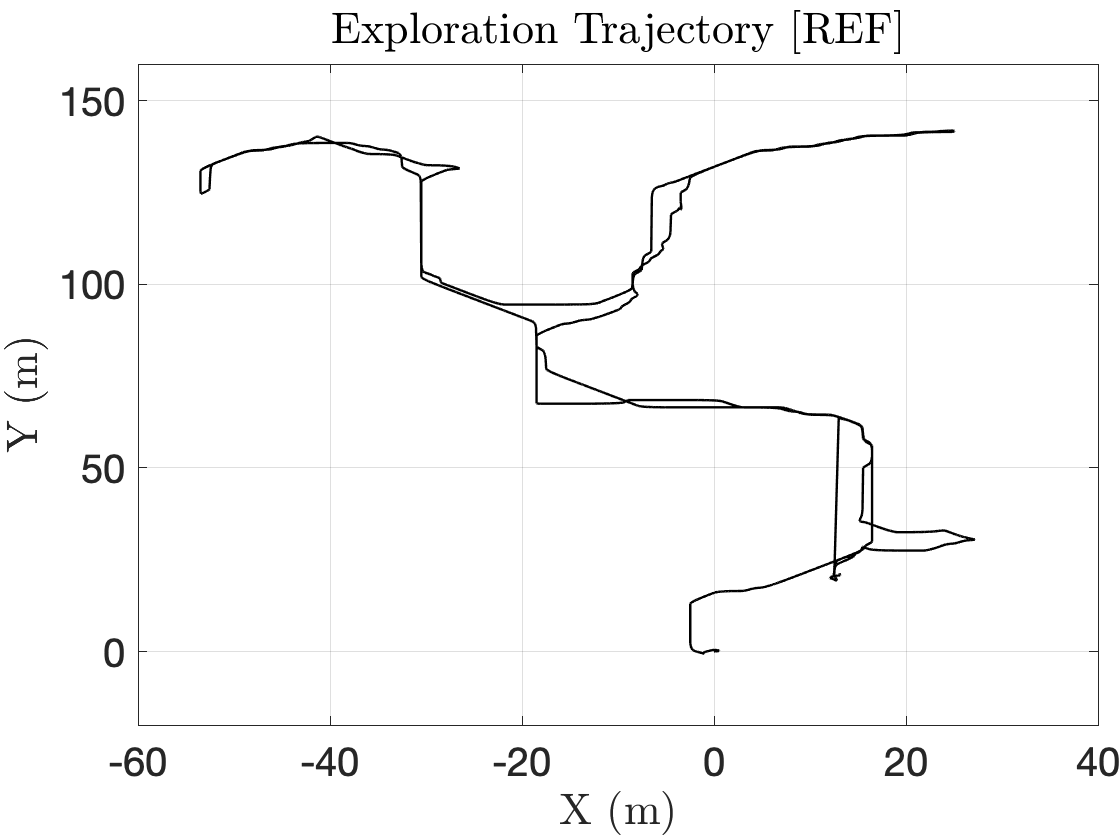}
        \label{fig:one}
        
    }
    \subfigure[]
    {
        \includegraphics[width=0.45\linewidth]{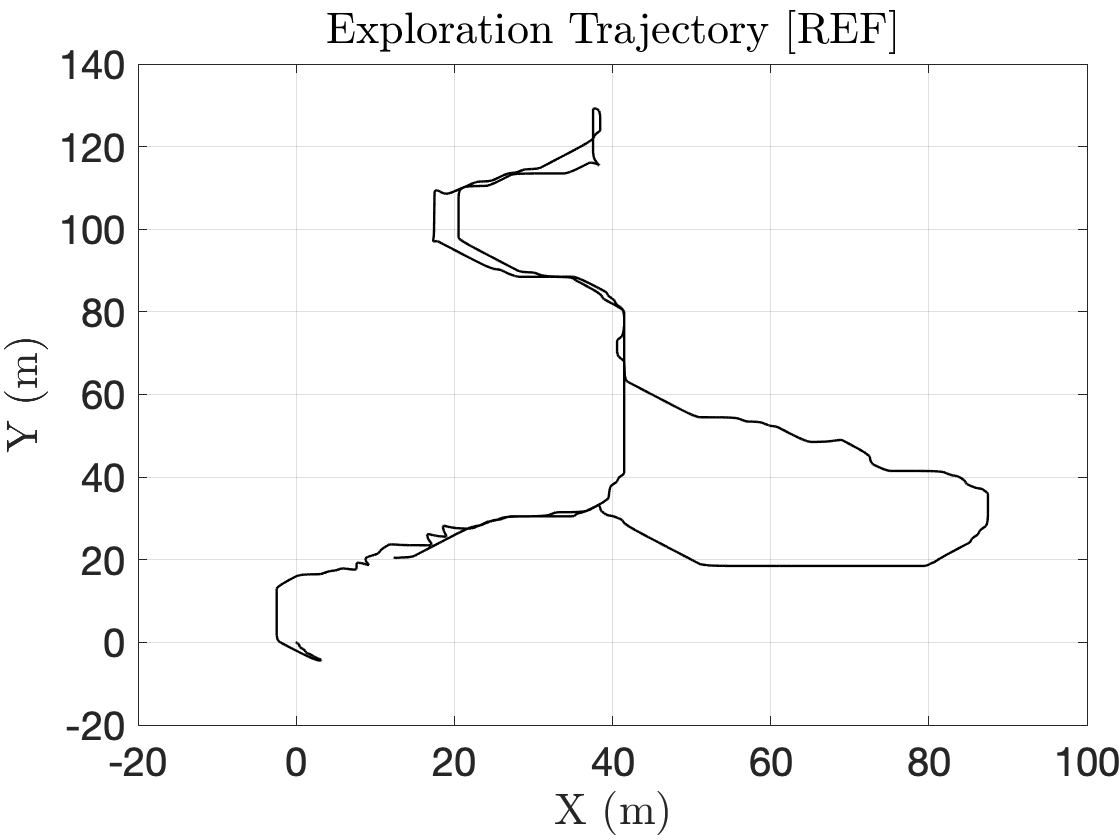}
        \label{fig:two}
        
    }
    \\
    \subfigure[]
    {
        \includegraphics[width=0.45\linewidth]{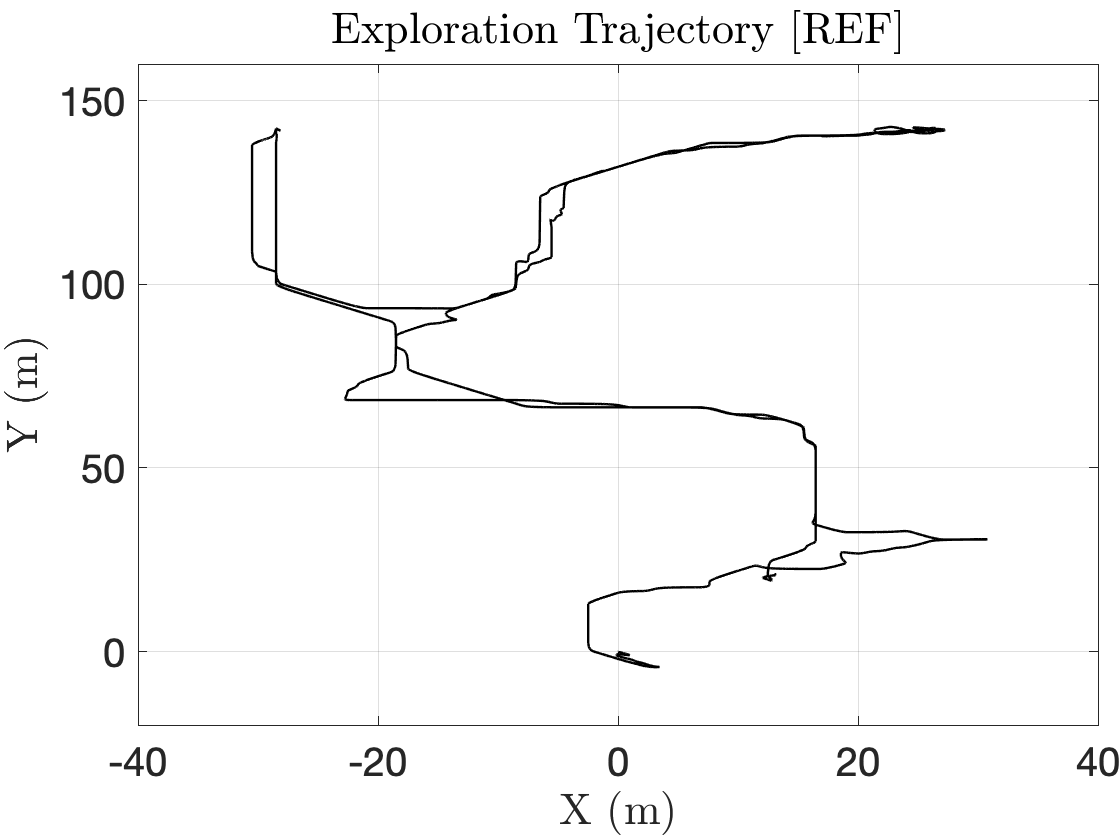}
        \label{fig:three}
        
    }
        \subfigure[]
    {
        \includegraphics[width=0.45\linewidth]{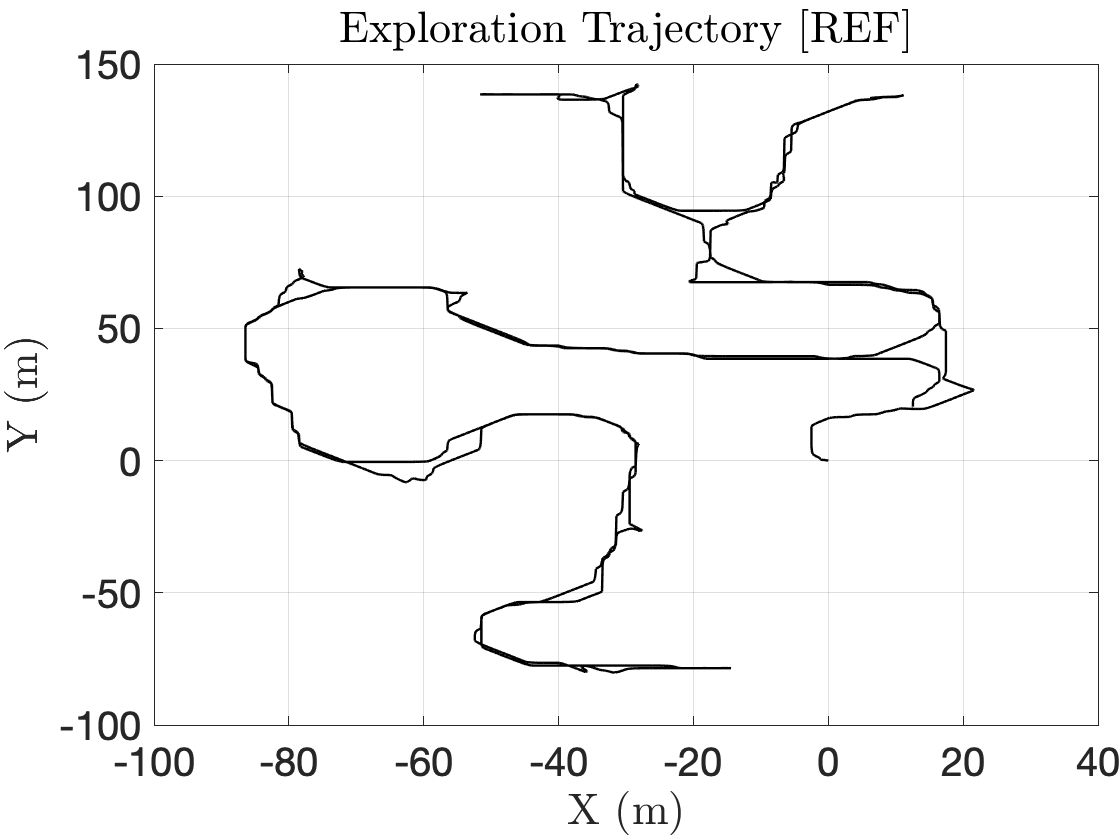}
        \label{fig:four}
        
    }
    \caption{REF equipped MAV explores a large and wide virtual cave environment with different mission duration and octree resolutions.}
    \label{fig:ourtrajs}
\end{figure*}

\begin{figure*}[h!]
    \centering
    \subfigure[]
    {
        \includegraphics[width=0.45\linewidth]{figures/trajthree.png}
        \label{fig:one}
        
    }
    \subfigure[]
    {
        \includegraphics[width=0.45\linewidth]{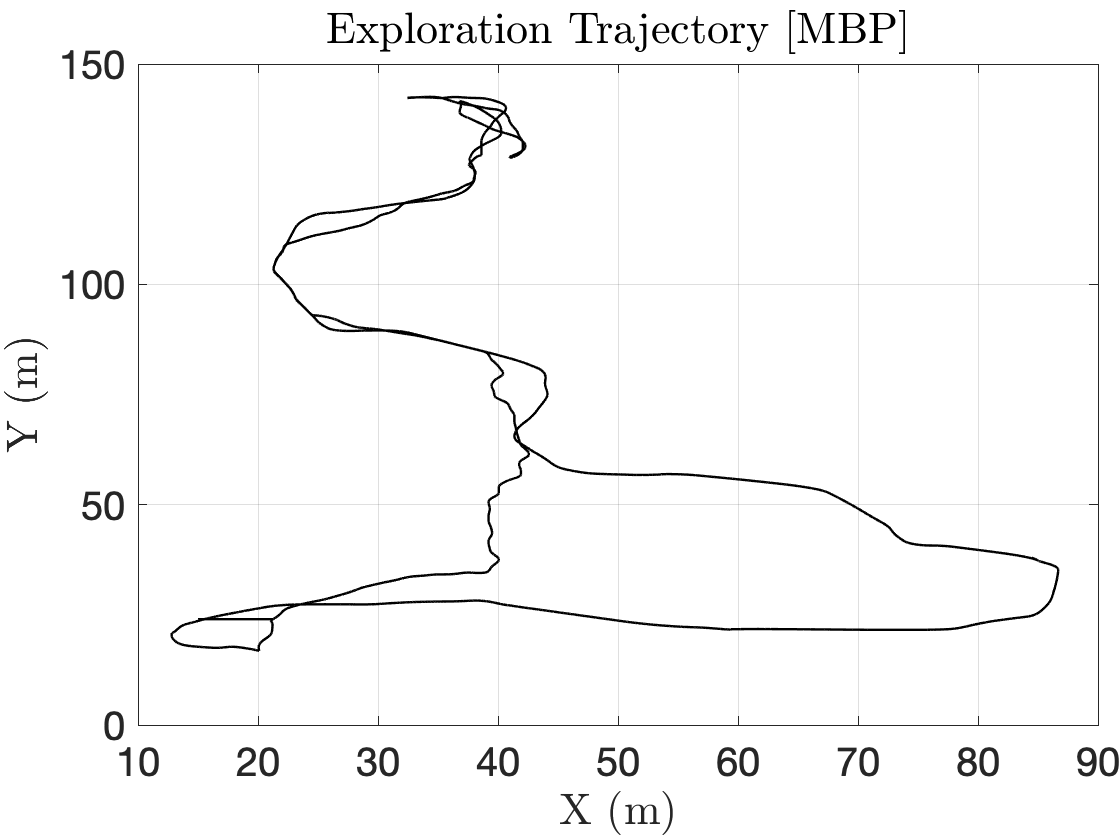}
        \label{fig:two}
        
    }
    \caption{400 sec mission: exploration trajectories, REF vs MB Planner. The proposed framework (REF) covers more ground in given time while avoiding loops in one area due to the global re-positioning functionality}
    \label{fig:comparetrajs}
\end{figure*}

In \autoref{fig:comparetrajs} the exploration mission trajectories are shown for REF and mbplanner with same mission duration (400 s). In \autoref{fig:comparetrajs} it is evident that the MAV covers more ground in given time using the proposed framework.  

In \autoref{fig:explorationtrajectory} in both method the overlap in trajectory is seen. This overlap is mainly due to the lower information gain (corresponding to mbplanner) and $\{\mathcal{L}\} = \varnothing$ (corresponding to our approach) resulting in the MAV to change direction and move to other unexplored areas. In \autoref{fig:Ours : 10min} it is evident that using the proposed global frontier selection strategy, the $NBF \in \{\mathcal{G}\}$ is selected such that the overlap in trajectory is minimal. In \autoref{fig:ourtrajs}, the MAV trajectory is tracked in $XY$ while exploring the lava tube virtual environment. The tracked trajectory is presented for visualizing the \textit{Look-Ahead-Move-Forward} nature of the proposed exploration framework. Due to such nature of exploration, the proposed framework is able to efficiently map new areas within given time and thus efficiently utilizing the resource constrained MAV's flight time. In Figure \ref{fig:one}, Figure \ref{fig:two} and Figure \ref{fig:three} a 400 second exploration mission is performed with different voxel resolution. In Figure \ref{fig:one}, Figure \ref{fig:two} and Figure \ref{fig:three} the exploration is performed with voxel resolution 0.3, 0.5 and 0.7 $m$ respectively. It is evident that corresponding to each voxel resolution in exploration mission, the MAV takes a different path while exploring based on the selected $NBF$ in each iteration. All exploration missions are performed with maximum forward velocity of the MAV as $1.5\ m/s$. In order to map the same environment even more quickly, an exploration mission with voxel resolution 0.9 m and mission duration 900 seconds is performed and the tracked trajectory of the MAV is presented in Figure \ref{fig:four}.



\begin{figure}[h!]
  \centering
  \captionsetup{justification=centering}
    \includegraphics[width=\linewidth]{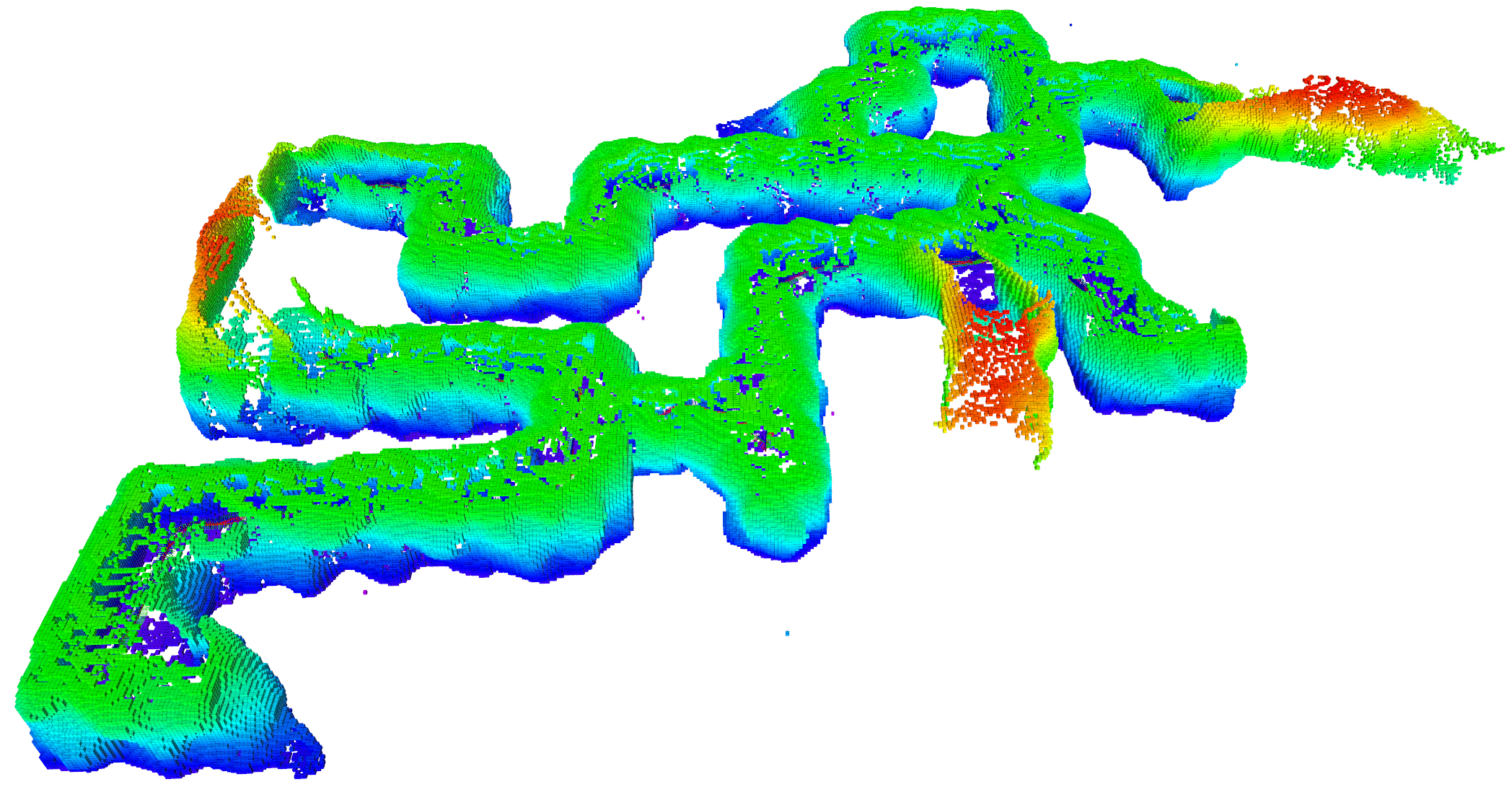}
    \caption{Octomap of the explored area using the proposed framework}
  \label{fig:octomap}
\end{figure}

\begin{figure}[h!]
  \centering
  \captionsetup{justification=centering}
    \includegraphics[width=\linewidth]{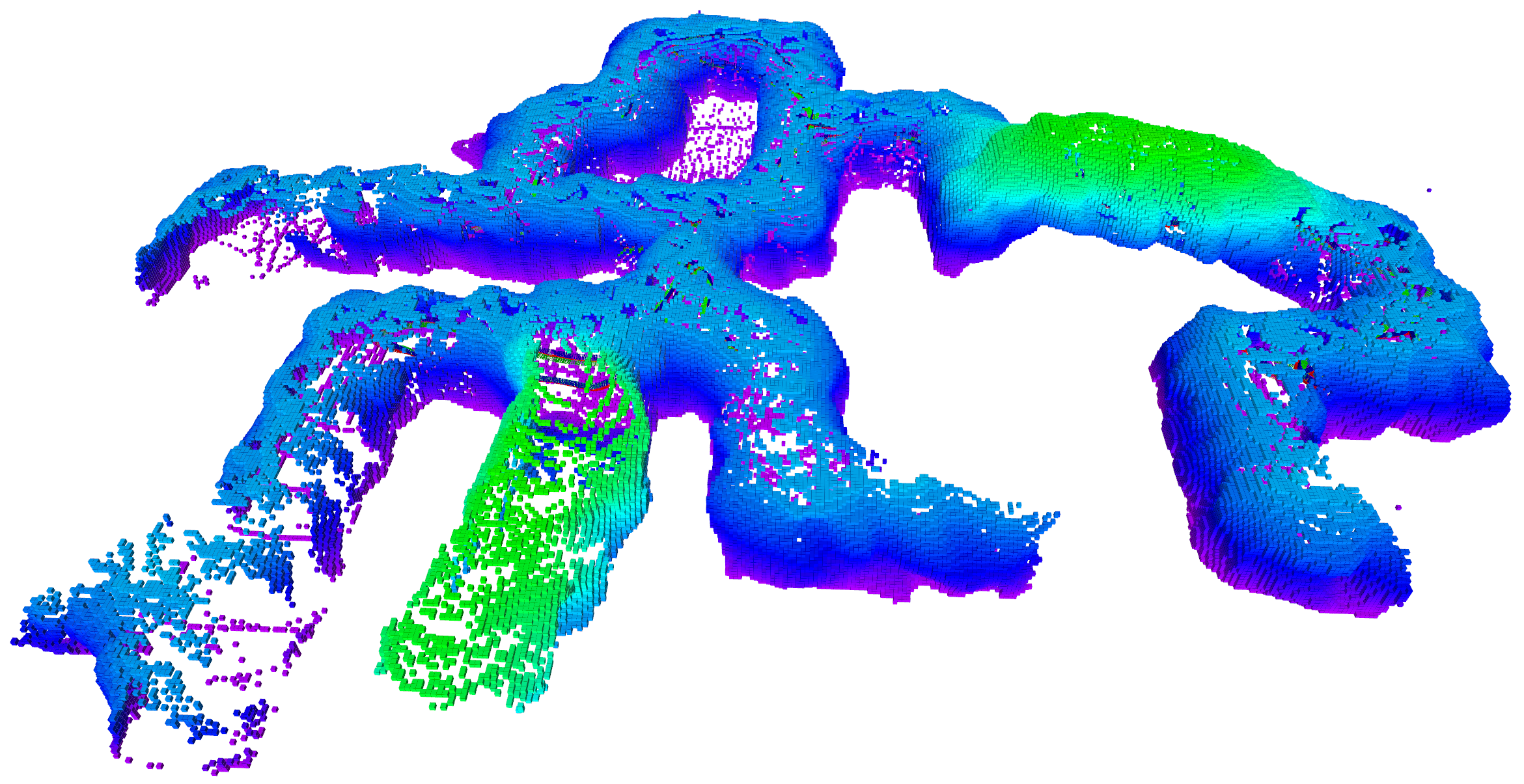}
    \caption{Octomap of the explored area using the mbplanner}
  \label{fig:octomapmbp}
\end{figure}

\section{Conclusions} \label{sec:conclusions}
In this article we proposed a Rapid Exploration Framework for deploying autonomous MAVs in unknown areas such as caves and mines. We present a novel candidate goal selection method with the focus of minimizing the actuation effort of the MAV by employing Look Forward Move Ahead approach. We compare the exploration scenario in the same environment with the motion primitive based planner which is a remarkable extension to Next Best View approach. In terms of volumetric gain and distance travelled, we achieve similar results to that of the mbplanner. We also address the trajectory overlap issue by introducing a simple yet efficient cost based goal selection approach that prevents the MAV to Unnecessarily travel to previously visited areas while also keeping the look forward move ahead approach as priority. As future development efforts we plan to conduct some field experiments to explore abandoned mines and underground cave structures.  

\section*{Declarations}

\textbf{Funding}: This work has been partially funded by the European Unions Horizon 2020 Research and Innovation Programme under the Grant Agreement No. 869379 illuMINEation.\\
\textbf{Conflict of interest}: The authors have no conflicts of interest with any related parties.\\
\textbf{Competing Interests}: Not applicable \\
\textbf{Code or data availability:} ROSbags of output data from simulated experiments can be made available at the suggestion of the reviewers and editors. \\
\textbf{Authors' Contributions}: Akash Patel: Development, implementation and system integration, relating to all presented submodules and developments, main manuscript contributors. Björn Lindqvist: Control and obstacle avoidance modules advisory. Christoforos Kanellakis: Software integration and High level advisory. Ali-akbar Agha-mohammadi: Advisory, development lead for Team CoSTAR in DARPA SubT Challenge. George Nikolakopoulos: Advisory, manuscript contributions, head of Luleå University of Technology Robotics\&AI Team. All authors have read and approved the manuscript.\\
\textbf{Ethics approval}: Not applicable.\\
\textbf{Consent to Participate}: Not applicable.\\
\textbf{Consent to Publish}: All authors comply with the consent to publish.\\

\bibliography{mybib}

\end{document}